# Formula Derivation and Analysis of the VINS-Mono

Yibin Wu

## I. INTRODUCTION

Simultaneous localization and mapping (SLAM) [1, 2] is considered as an essential technology for autonomous mobile robots to effectively and safely operate in unknown environments. SLAM consists in the concurrent construction of a model of the environment (the map), and the estimation of the state of the robot moving within it [3]. Visual odometry (VO) ca be considered a reduced SLAM system, in which the loop closure (or place recognition) module disabled [3]. Monocular vision-based VO is incapable of recovering the metric scale and suffers from illumination change, less texture, and etc. Therefore, inertial measurement unit (IMU) has been extensively introduced to integrate with vision system to construct the visual-inertial SLAM (VI-SLAM) or visual-inertial odometry (VIO). In the last decades, a large amount of research have significantly promoted the development and applications of the visual or visual-inertial system. A recent survey of the current state of SLAM can be found in [3]. A summary of the most representative visual and visual-inertial systems is listed in [4]

The VINS-Mono [5, 6] is a monocular visual-inertial 6 DOF state estimator proposed by Aerial Robotics Group of HKUST in 2017. It can be performed on MAVs, smartphones and many other intelligent platforms. Because of the significant robustness, accuracy and scalability, it has gained extensive attention worldwide. The whole framework of the VINS-Mono is shown as Fig. 1. The main contribution includes:

1. A robust initialization procedure which can produce a comparatively accurate estimation of the visual scale, gravity, velocity and gyroscope bias.
2. Sliding window-based local optimization.
3. Online relocalization and 4 DOF global pose graph optimization.

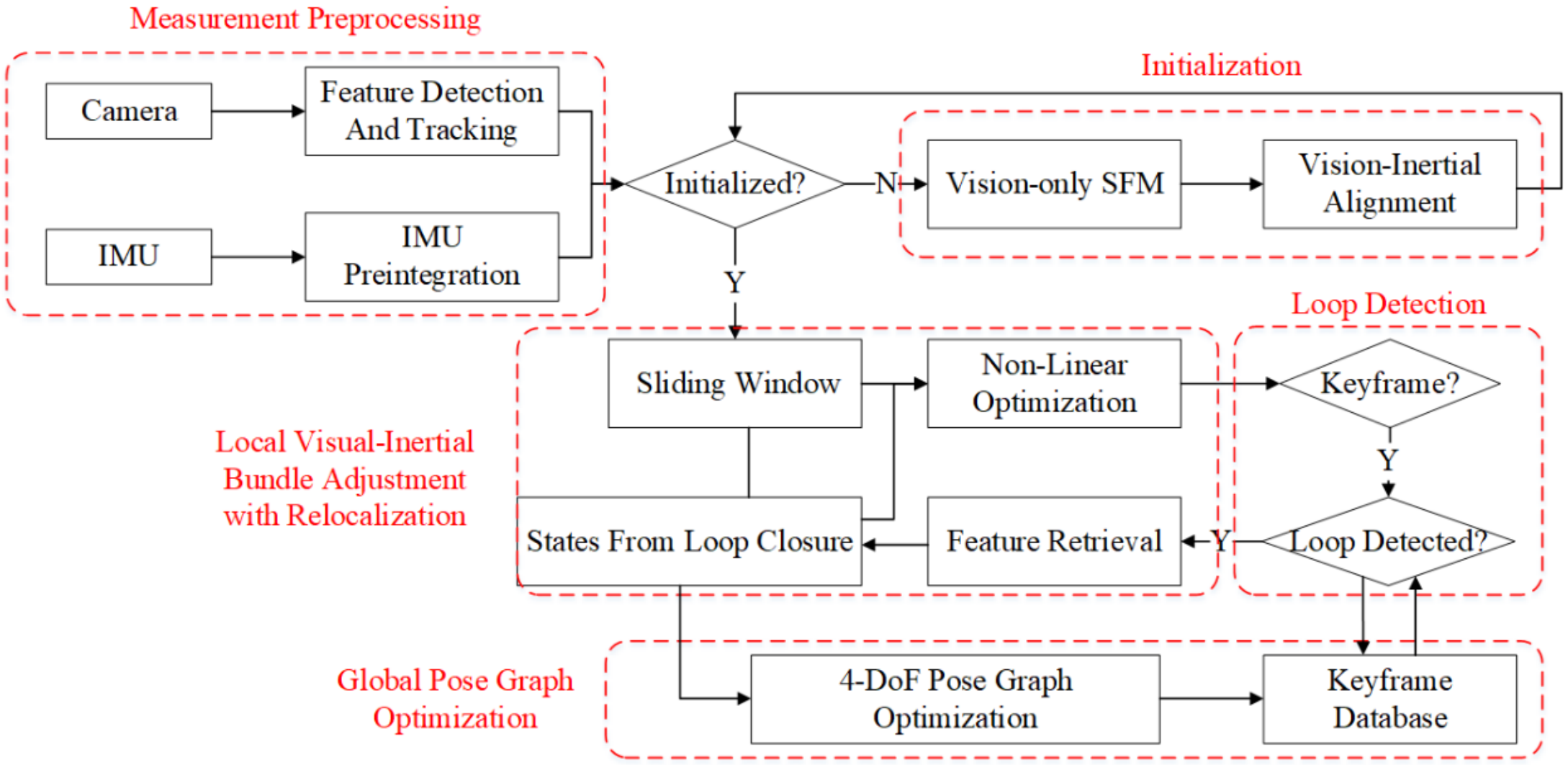

Fig. 1 Framework overview of the VINS-Mono (Shaojie Shen, 2017).

This manuscript is organized as following. The process of the IMU preintegration is described in Section II. Section III presents the visual/inertial co-initialization procedure. The tightly-coupled nonlinear optimization is then derived in Section IV. Section V deduces the marginalization. The derivation of the global optimization with GPS in VINS-Fusion [7], which is an extension of VIN-Mono, are provided in Section VI. Refer to the Appendix for details of some equation derivation.

**NOMENCLATURE**

a) Matrices are denoted as upper case bold letters.
b) Vectors are denoted as lower case bold italic letters.
c) Scalar is denoted as lower case italic letters.
d) The coordinate frames involved in the vector transformation are denoted as superscript and subscript. For vectors, the superscript denotes the projected coordinate system.
e) $\hat{*}$, estimated or computed values.
f) $\tilde{*}$, observed or measured values.
g) $\boldsymbol{a}_x$, element of vector $\boldsymbol{a}$ on $x$ axis.

## II. IMU PREINTEGRATION

### *2.1 IMU Preintegration in Continuous Time*

The IMU preintegration is proposed in [8, 9]. In most practical applications, the IMU data rate is larger than that of the camera, which means that there are usually several IMU measurements between two consecutive frames, as shown in Fig. 2. First of all, we have to know that an IMU consists of a gyroscope and an accelerometer to measure the angular rate and acceleration of the IMU w.r.t. the inertial frame, respectively. For accelerometer, its measurement can be written as

$$\boldsymbol{f}^b = \boldsymbol{a}^b - \boldsymbol{g}^b \tag{1}$$

where $\boldsymbol{f}$ is the special force, $\boldsymbol{\alpha}$ is the true acceleration of the IMU, $\boldsymbol{g}$ is the local gravity. It means that the output of the accelerometer is not the true acceleration of the IMU, but the acceleration minus gravity. For example, if the IMU frame is defined as right-front-up, its measurement will be $\boldsymbol{f}_0^b = -\boldsymbol{g}^b \approx 9.8m/s$ when the IMU keeps static and level.

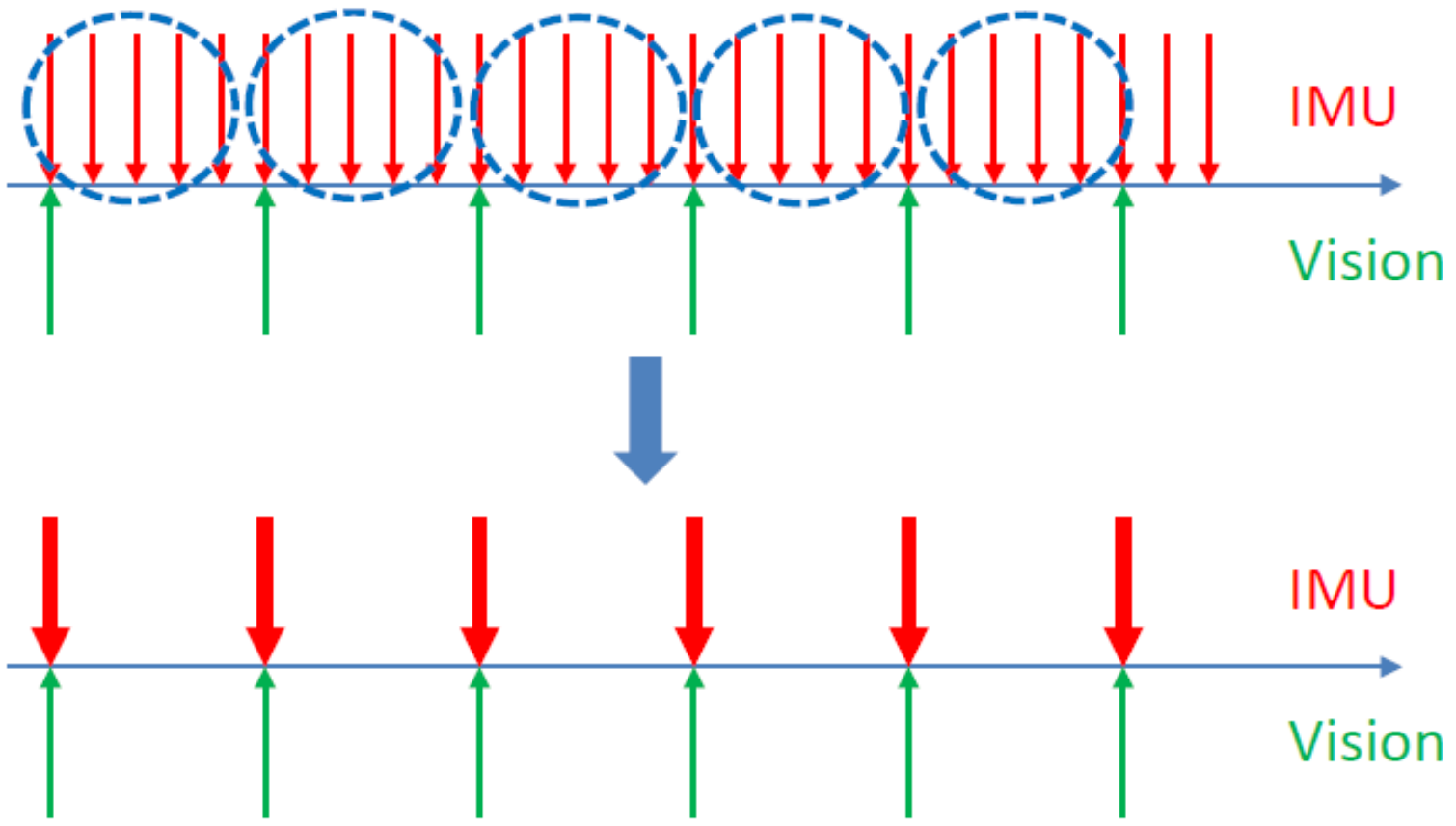

Fig. 2 Illustration of the IMU preintegration (Shaojie Shen, 2017).

Given two time instants that correspond to image frames $b_k$ and $b_{k+1}$, the state variables are constrained by inertial measurements during time interval $[k, k+1]$:

$$\begin{aligned}
\boldsymbol{p}_{b_{k+1}}^{w} &= \boldsymbol{p}_{b_k}^{w} + \boldsymbol{v}_{b_k}^{w} \Delta t_k + \iint_{t\in[k,k+1]} (\mathbf{R}_t^{w}(\hat{\boldsymbol{a}}_t - \boldsymbol{b}_{at}) - \boldsymbol{g}^{w}) dt^2 \\
\boldsymbol{v}_{b_{k+1}}^{w} &= \boldsymbol{v}_{b_k}^{w} + \int_{t\in[k,k+1]} (\mathbf{R}_t^{w}(\hat{\boldsymbol{a}}_t - \boldsymbol{b}_{at}) - \boldsymbol{g}^{w}) dt \\
\boldsymbol{q}_{b_{k+1}}^{w} &= \boldsymbol{q}_{b_k}^{w} \otimes \int_{t\in[k,k+1]} \frac{1}{2} \boldsymbol{\Omega}(\hat{\boldsymbol{\omega}}_t - \boldsymbol{b}_{gt}) \boldsymbol{q}_t^{b_k} dt
\end{aligned} \tag{2}$$

**NOTICE**: Here, $\boldsymbol{g}^w$ does not denote the gravity in world coordinates but the projection of $\boldsymbol{f}_0^b$ in the world frame.

Where

$$\boldsymbol{\Omega}_R(\boldsymbol{\omega}) = [\boldsymbol{\omega}]_{\mathrm{R}} = \begin{bmatrix} 0 & -\boldsymbol{\omega}^{\mathrm{T}} \\ \boldsymbol{\omega} & (\boldsymbol{\omega})\times \end{bmatrix}, (\boldsymbol{\omega})\times = \begin{bmatrix} 0 & -\omega_z & \omega_y \\ \omega_z & 0 & -\omega_x \\ -\omega_y & \omega_x & 0 \end{bmatrix} \tag{3}$$

For the last equation in (2), according to [10], we have

$$\begin{aligned}
\dot{\boldsymbol{q}} &\triangleq \lim_{\delta t\to 0} \frac{\boldsymbol{q}(t+\delta t) - \boldsymbol{q}(t)}{\delta t} \\
&= \lim_{\delta t\to 0} \frac{\boldsymbol{q}(t) \otimes \delta \boldsymbol{q} - \boldsymbol{q}(t)}{\delta t} \\
&= \lim_{\delta t\to 0} \frac{\boldsymbol{q}(t) \otimes \left( \begin{bmatrix} 1 \\ \delta\boldsymbol{\theta}/2 \end{bmatrix} - \begin{bmatrix} 1 \\ 0 \end{bmatrix} \right)}{\delta t} \\
&= \frac{1}{2} \boldsymbol{q} \otimes \boldsymbol{\omega}_t = \frac{1}{2} \boldsymbol{\Omega}_R(\boldsymbol{\omega}_t) \boldsymbol{q}
\end{aligned} \tag{4}$$

In this manuscript, we define that $\boldsymbol{q} = \begin{bmatrix} q_w & \boldsymbol{q}_{\mathbf{v}} \end{bmatrix}^{\mathrm{T}} = \begin{bmatrix} q_w & q_x & q_y & q_z \end{bmatrix}^{\mathrm{T}}$.

Changing the reference frame $w$ of IMU propagation to local frame $b_k$, we can only integrate the parts which are related to linear acceleration $\hat{\boldsymbol{a}}$ and angular velocity $\hat{\boldsymbol{\omega}}$ as follows:

$$\begin{aligned}
\mathbf{R}_w^{b_k} \boldsymbol{p}_{b_{k+1}}^w &= \mathbf{R}_w^{b_k} (\boldsymbol{p}_{b_k}^w + \boldsymbol{v}_{b_k}^w \Delta t_k - \frac{1}{2} \boldsymbol{g}^w \Delta t_k^{\ 2}) + \boldsymbol{\alpha}_{b_{k+1}}^{b_k} \\
\mathbf{R}_w^{b_k} \boldsymbol{v}_{b_{k+1}}^w &= \mathbf{R}_w^{b_k} (\boldsymbol{v}_{b_k}^w - \boldsymbol{g}^w \Delta t_k) + \boldsymbol{\beta}_{b_{k+1}}^{b_k} \\
\boldsymbol{q}_w^{b_k} \boldsymbol{q}_{b_{k+1}}^w &= \boldsymbol{\gamma}_{b_{k+1}}^{b_k}
\end{aligned} \tag{5}$$

where

$$\begin{aligned}
\boldsymbol{\alpha}_{b_{k+1}}^{b_k} &= \iint_{t \in [k,k+1]} \mathbf{R}_t^{b_k} (\hat{\boldsymbol{a}}_t - \boldsymbol{b}_{at})) dt^2 \\
\boldsymbol{\beta}_{b_{k+1}}^{b_k} &= \int_{t \in [k,k+1]} \mathbf{R}_t^{b_k} (\hat{\boldsymbol{a}}_t - \boldsymbol{b}_{at}) dt \\
\boldsymbol{\gamma}_{b_{k+1}}^{b_k} &= \int_{t \in [k,k+1]} \frac{1}{2} \boldsymbol{\Omega}(\hat{\boldsymbol{\omega}}_t - \boldsymbol{b}_{gt}) \boldsymbol{q}_t^{b_k} dt
\end{aligned} \tag{6}$$

*2.2 IMU Preintegration in Discrete Time*

As for the discrete-time implementation of the IMU preintegration, we can apply Mid-point method as following

$$\begin{aligned}
\boldsymbol{\gamma}_{i+1}^{b_k} &= \boldsymbol{\gamma}_i^{b_k} \otimes \boldsymbol{\gamma}_{i,i+1}^{b_k} = \boldsymbol{\gamma}_i^{b_k} \otimes \begin{bmatrix} 1 \\ 1/2(\boldsymbol{\omega}_i + \boldsymbol{\omega}_{i+1} - 2\boldsymbol{b}_{gi})\delta t \end{bmatrix} \\
\boldsymbol{\beta}_{i+1}^{b_k} &= \int_{t \in [k,i]} (\ldots) dt + \int_{t \in [i,i+1]} (\ldots) dt \\
&= \boldsymbol{\beta}_i^{b_k} + \frac{1}{2} \left[ \mathbf{R}(\boldsymbol{\gamma}_i^{b_k})(\hat{\boldsymbol{a}}_i - \boldsymbol{b}_{ai}) + \mathbf{R}(\boldsymbol{\gamma}_{i+1}^{b_k})(\hat{\boldsymbol{a}}_{i+1} - \boldsymbol{b}_{ai}) \right] \delta t \\
\boldsymbol{\alpha}_{i+1}^{b_k} &= \boldsymbol{\alpha}_i^{b_k} + \frac{1}{2} (\boldsymbol{\beta}_i^{b_k} + \boldsymbol{\beta}_{i+1}^{b_k}) \delta t \\
&= \boldsymbol{\alpha}_i^{b_k} + \boldsymbol{\beta}_i^{b_k} \delta t + \frac{1}{4} \left[ \mathbf{R}(\boldsymbol{\gamma}_i^{b_k})(\hat{\boldsymbol{a}}_i - \boldsymbol{b}_{ai}) + \mathbf{R}(\boldsymbol{\gamma}_{i+1}^{b_k})(\hat{\boldsymbol{a}}_{i+1} - \boldsymbol{b}_{ai}) \right] \delta t^2
\end{aligned} \tag{7}$$

**NOTICE:** Strictly speaking, the preintegration values of position $\boldsymbol{\alpha}_{b_{k+1}}^{b_k}$ and velocity $\boldsymbol{\beta}_{b_{k+1}}^{b_k}$ do not have any physical meaning, since the gravity is ignored in the integral process. However, we can imagine a zero gravity space in which the measurement of accelerometer will be the real acceleration, then Eq. (6) and (7) become easy to understand.

*2.3 Error-state Kinematics in Continuous Time*

Inspired by [10], we can write the error-state equations of the kinematics of an inertial system. We introduce the error perturbation analysis as following

$$\hat{\boldsymbol{x}} = \boldsymbol{x} + \delta \boldsymbol{x} \tag{8}$$

where $\boldsymbol{x}$ is the ideal (or nominal [10]) value of the sensor measurements or the body state without any error; $\hat{\boldsymbol{x}}$ is the calculated (or observed, or true [10]) value of measurements and body state which contain errors $\delta \boldsymbol{x}$.

For the ideal-state kinematics, we have

$$\begin{aligned} &\dot{\boldsymbol{\alpha}} = \boldsymbol{\beta} \\ &\dot{\boldsymbol{\beta}} = \mathbf{R}(\boldsymbol{a}_m - \boldsymbol{b}_a) \\ &\dot{\gamma} = \frac{1}{2}\gamma \otimes (\boldsymbol{\omega}_m - \boldsymbol{b}_\omega) \\ &\dot{\boldsymbol{b}}_a = 0 \\ &\dot{\boldsymbol{b}}_\omega = 0 \end{aligned} \tag{9}$$

For the calculated-state, we have

$$\begin{aligned} &\dot{\hat{\boldsymbol{\alpha}}} = \hat{\boldsymbol{\beta}} \\ &\dot{\hat{\boldsymbol{\beta}}} = \hat{\mathbf{R}}(\boldsymbol{a}_m - \boldsymbol{n}_a - \hat{\boldsymbol{b}}_a) \\ &\dot{\hat{\gamma}} = \frac{1}{2}\hat{\gamma} \otimes \left(\boldsymbol{\omega}_m - \boldsymbol{n}_\omega - \hat{\boldsymbol{b}}_g\right) \\ &\dot{\hat{\boldsymbol{b}}}_a = \boldsymbol{n}_{b_a} \\ &\dot{\hat{\boldsymbol{b}}}_\omega = \boldsymbol{n}_{b_\omega} \end{aligned} \tag{10}$$

Assume that the acceleration bias and gyroscope bias are random walk, whose derivatives are Gaussian white noise, i.e., $n_{b_a} \sim N(0, \sigma_{b_a}^2), n_{b_\omega} \sim N(0, \sigma_{b_\omega}^2)$. The noise in acceleration $\boldsymbol{a}_m$ and gyroscope measurements $\boldsymbol{\omega}_m$ are treated as the same.

Then, we have the error-state kinematics,

$$\begin{aligned} &\dot{\delta\boldsymbol{\alpha}} = \delta\boldsymbol{\beta} \\ &\dot{\delta\boldsymbol{\beta}} = -\mathbf{R}(\boldsymbol{a}_m - \boldsymbol{b}_a) \times \delta\boldsymbol{\theta} - \mathbf{R}\boldsymbol{n}_a - \mathbf{R}\delta\boldsymbol{b}_a \\ &\dot{\delta\boldsymbol{\theta}} = -(\boldsymbol{\omega}_m - \boldsymbol{b}_g) \times \delta\boldsymbol{\theta} - \delta\boldsymbol{b}_g - \boldsymbol{n}_\omega \\ &\dot{\delta\boldsymbol{b}}_a = \boldsymbol{n}_{b_a} \\ &\dot{\delta\boldsymbol{b}}_g = \boldsymbol{n}_{b_g} \end{aligned} \tag{11}$$

Derivations of the differential equation of the velocity and attitude errors are developed as follows, the second-order small terms are so trivial that we ignore it,

$$
\begin{aligned}
\dot{\boldsymbol{\beta}}+\delta\dot{\boldsymbol{\beta}} &= \dot{\hat{\boldsymbol{\beta}}} = \hat{\mathbf{R}}(\boldsymbol{a}_m-\boldsymbol{n}_a-\hat{\boldsymbol{b}}_a) \\
\mathbf{R}(\boldsymbol{a}_m-\boldsymbol{b}_a)+\delta\dot{\boldsymbol{\beta}} &= \dot{\hat{\boldsymbol{\beta}}} = \mathbf{R}[\mathbf{I}+(\delta\boldsymbol{\theta})\times](\boldsymbol{a}_m-\boldsymbol{n}_a-\boldsymbol{b}_a-\delta\boldsymbol{b}_a) \\
\delta\dot{\boldsymbol{\beta}} &= -\mathbf{R}(\boldsymbol{a}_m-\boldsymbol{b}_a)\times\delta\boldsymbol{\theta}-\mathbf{R}\boldsymbol{n}_a-\mathbf{R}\delta\boldsymbol{b}_a
\end{aligned}
\tag{12}
$$

$$
\begin{aligned}
&\gamma\otimes\dot{\delta\gamma} = \dot{\hat{\gamma}} = \frac{1}{2}\hat{\gamma}\otimes\left(\boldsymbol{\omega}_m-\boldsymbol{n}_\omega-\hat{\boldsymbol{b}}_g\right) \\
&\dot{\gamma}\otimes\delta\gamma+\gamma\otimes\dot{\delta\gamma} = \dot{\hat{\gamma}} = \frac{1}{2}\gamma\otimes\delta\gamma\otimes\left(\boldsymbol{\omega}_m-\boldsymbol{n}_\omega-\boldsymbol{b}_g-\delta\boldsymbol{b}_\omega\right) \\
&\frac{1}{2}\gamma\otimes\left(\boldsymbol{\omega}_m-\boldsymbol{b}_g\right)\otimes\delta\gamma+\gamma\otimes\dot{\delta\gamma} = \frac{1}{2}\gamma\otimes\delta\gamma\otimes\left(\boldsymbol{\omega}_m-\boldsymbol{n}_\omega-\boldsymbol{b}_g-\delta\boldsymbol{b}_g\right) \\
&let(\hat{\boldsymbol{\omega}}=\boldsymbol{\omega}_m-\boldsymbol{n}_\omega-\boldsymbol{b}_g-\delta\boldsymbol{b}_g,\boldsymbol{\omega}=\boldsymbol{\omega}_m-\boldsymbol{b}_g) \\
&2\dot{\delta\gamma} = \delta\gamma\otimes\hat{\boldsymbol{\omega}}-\boldsymbol{\omega}\otimes\delta\gamma \\
&= \boldsymbol{\Omega}_R(\hat{\boldsymbol{\omega}})\delta\gamma-\boldsymbol{\Omega}_L(\boldsymbol{\omega})\delta\gamma \\
&= \begin{bmatrix} 0 & (\boldsymbol{\omega}-\hat{\boldsymbol{\omega}})^{\mathrm{T}} \\ \hat{\boldsymbol{\omega}}-\boldsymbol{\omega} & -(\boldsymbol{\omega}+\hat{\boldsymbol{\omega}})\times \end{bmatrix}\delta\gamma \\
&\begin{bmatrix} 0 \\ \delta\dot{\boldsymbol{\theta}} \end{bmatrix} = \begin{bmatrix} 0 & \left(\delta\boldsymbol{b}_g+\boldsymbol{n}_\omega\right)^{\mathrm{T}} \\ -\delta\boldsymbol{b}_g-\boldsymbol{n}_\omega & -\left(2\boldsymbol{\omega}_m-\boldsymbol{n}_\omega-2\boldsymbol{b}_g-\delta\boldsymbol{b}_g\right)\times \end{bmatrix}\begin{bmatrix} 1 \\ \frac{1}{2}\delta\boldsymbol{\theta} \end{bmatrix} \\
&\delta\dot{\boldsymbol{\theta}} = -(\boldsymbol{\omega}_m-\boldsymbol{b}_g)\times\delta\boldsymbol{\theta}-\delta\boldsymbol{b}_g-\boldsymbol{n}_\omega
\end{aligned}
\tag{13}
$$

Then, we have

$$
\begin{bmatrix} \delta\dot{\boldsymbol{\alpha}}_t^{b_k} \\ \delta\dot{\boldsymbol{\beta}}_t^{b_k} \\ \delta\dot{\boldsymbol{\theta}}_t^{b_k} \\ \delta\dot{\boldsymbol{b}}_{a_t} \\ \delta\dot{\boldsymbol{b}}_{g_t} \end{bmatrix} = \begin{bmatrix} & \mathbf{I} & & & \\ & & -\mathbf{R}_t^{b_k}(\boldsymbol{a}_m-\boldsymbol{b}_a)\times & -\mathbf{R}_t^{b_k} & \\ & & -(\boldsymbol{\omega}_m-\boldsymbol{b}_g)\times & & -\mathbf{I} \\ & & & & \\ & & & & \end{bmatrix}\begin{bmatrix} \delta\boldsymbol{\alpha}_t^{b_k} \\ \delta\boldsymbol{\beta}_t^{b_k} \\ \delta\boldsymbol{\theta}_t^{b_k} \\ \delta\boldsymbol{b}_{a_t} \\ \delta\boldsymbol{b}_{\omega_t} \end{bmatrix} + \begin{bmatrix} & & & \\ -\mathbf{R}_t^{b_k} & & & \\ & -\mathbf{I} & & \\ & & \mathbf{I} & \\ & & & \mathbf{I} \end{bmatrix}\begin{bmatrix} \boldsymbol{n}_a \\ \boldsymbol{n}_\omega \\ \boldsymbol{n}_{b_a} \\ \boldsymbol{n}_{b_\omega} \end{bmatrix}
\tag{14}
$$

$$
\delta\dot{\boldsymbol{x}}_t = \mathbf{F}_t\delta\boldsymbol{x}_t+\mathbf{G}_t\boldsymbol{n}_t
\tag{15}
$$

As we all know,

$$
\dot{\boldsymbol{x}} = \lim_{\delta t\to\infty}\frac{\boldsymbol{x}\left(t+\delta t\right)-\boldsymbol{x}\left(t\right)}{\delta t}
\tag{16}
$$

Within a small time interval, we have

$$
\delta\boldsymbol{x}_{t+\delta t} = \left(\mathbf{I}+\mathbf{F}_t\delta t\right)\delta\boldsymbol{x}_t+\mathbf{G}_t\boldsymbol{n}_t\delta t
\tag{17}
$$

$\mathbf{P}_{b_{k+1}}^{b_k}$ can be computed recursively by the first-order discrete-time covariance updating with the initial covariance $\mathbf{P}_{b_k}^{b_k} = 0$:

$$\mathbf{P}_{t+\delta t}^{b_k} = \left(\mathbf{I}+\mathbf{F}_t\delta t\right)\mathbf{P}_t^{b_k}\left(\mathbf{I}+\mathbf{F}_t\delta t\right)^{\mathrm{T}} + \left(\mathbf{G}_t\delta t\right)\mathbf{Q}\left(\mathbf{G}_t\delta t\right)^{\mathrm{T}} \tag{18}$$

Where $\delta t$ is the time between two IMU measurements, and $\mathbf{Q}$ is the diagonal covariance matrix of noise $diag\left(\sigma_a^2, \sigma_\omega^2, \sigma_{b_a}^2, \sigma_{b_\omega}^2\right)$. Meanwhile, the first-order Jacobian matrix $\mathbf{J}_{b_{k+1}}^{b_k}$ of $\delta \boldsymbol{x}_{b_{k+1}}^{b_k}$ w.r.t $\delta \boldsymbol{x}_{b_k}^{b_k}$ can be compute recursively with the initial Jacobian $\mathbf{J}_{b_k}^{b_k} = \mathbf{I}$,

$$\mathbf{J}_{t+\delta t} = \left(\mathbf{I} + \mathbf{F}_t\delta t\right)\mathbf{J}_t, t \in [k, k+1] \tag{19}$$

Consider the variables associated with the preintegration, the first order approximation of $\left(\boldsymbol{\alpha}_{b_{k+1}}^{b_k},\ \boldsymbol{\beta}_{b_{k+1}}^{b_k},\ \boldsymbol{\gamma}_{b_{k+1}}^{b_k}\right)$ can be wrote as

$$\begin{aligned}
\boldsymbol{\alpha}_{b_{k+1}}^{b_k} &\approx \hat{\boldsymbol{\alpha}}_{b_{k+1}}^{b_k} + \mathbf{J}_{b_a}^{\alpha}\delta\boldsymbol{b}_{ak} + \mathbf{J}_{b_g}^{\alpha}\delta\boldsymbol{b}_{gk} \\
\boldsymbol{\beta}_{b_{k+1}}^{b_k} &\approx \hat{\boldsymbol{\beta}}_{b_{k+1}}^{b_k} + \mathbf{J}_{b_a}^{\beta}\delta\boldsymbol{b}_{ak} + \mathbf{J}_{b_g}^{\beta}\delta\boldsymbol{b}_{gk} \\
\boldsymbol{\gamma}_{b_{k+1}}^{b_k} &= \hat{\boldsymbol{\gamma}}_{b_{k+1}}^{b_k} \otimes \begin{bmatrix} 1 \\ \frac{1}{2}\mathbf{J}_{b_g}^{\gamma}\delta\boldsymbol{b}_{gk} \end{bmatrix}
\end{aligned} \tag{20}$$

where $\mathbf{J}_{b_a}^{\alpha}$ is the sub-block matrix in $\mathbf{J}_{b_{k+1}}$ whose location is corresponding to $\frac{\delta \boldsymbol{x}_{b_{k+1}}^{b_k}}{\delta \boldsymbol{b}_a}$, the others are of the similar meaning.

*2.4 Error-state Kinematics in Discrete Time*

We assume that the sensor biases between two IMU measurements are constant. According to Eq. (17) and using the Mid-point integration for discrete time implementation, we have

$$
\begin{cases}
\dot{\delta\boldsymbol{\theta}}_t^{b_k} = -(\frac{\boldsymbol{\omega}_t+\boldsymbol{\omega}_{t+1}}{2} - \boldsymbol{b}_{\omega_t})\times\delta\boldsymbol{\theta}_t^{b_k} - \delta\boldsymbol{b}_{g_t} - \frac{\boldsymbol{n}_{\omega_t}+\boldsymbol{n}_{\omega_{t+1}}}{2} \\
\delta\boldsymbol{\theta}_{t+1}^{b_k} = \left[\mathbf{I} - \delta t(\frac{\boldsymbol{\omega}_t+\boldsymbol{\omega}_{t+1}}{2} - \boldsymbol{b}_{g_t})\times\right]\delta\boldsymbol{\theta}_t^{b_k} - \delta\boldsymbol{b}_{g_t}\delta t - \frac{\boldsymbol{n}_{\omega_t}+\boldsymbol{n}_{\omega_{t+1}}}{2}\delta t
\end{cases}
$$

$$
\begin{cases}
\dot{\delta\boldsymbol{\beta}}_t^{b_k} = -\mathbf{R}_t^{b_k}(\boldsymbol{a}_t - \boldsymbol{b}_{a_t})\times\delta\boldsymbol{\theta}_t^{b_k} - \mathbf{R}_t^{b_k}\delta\boldsymbol{b}_{a_t} - \mathbf{R}_t^{b_k}\boldsymbol{n}_{a_t} \\
\quad = -\frac{1}{2}\mathbf{R}_t^{b_k}(\boldsymbol{a}_t - \boldsymbol{b}_{a_t})\times\delta\boldsymbol{\theta}_t^{b_k} - \frac{1}{2}\mathbf{R}_{t+1}^{b_k}(\boldsymbol{a}_{t+1} - \boldsymbol{b}_{a_t})\times\delta\boldsymbol{\theta}_{t+1}^{b_k} - \frac{1}{2}\left(\mathbf{R}_t^{b_k} + \mathbf{R}_{t+1}^{b_k}\right)\delta\boldsymbol{b}_{a_t} \\
\quad -\frac{1}{2}\left(\mathbf{R}_t^{b_k} + \mathbf{R}_{t+1}^{b_k}\right)\boldsymbol{n}_{a_t} \\
\delta\boldsymbol{\beta}_{t+1}^{b_k} = \left\{-\frac{\delta t}{2}\mathbf{R}_t^{b_k}\left(\boldsymbol{a}_t - \boldsymbol{b}_{a_t}\right)\times - \frac{\delta t^2}{2}\mathbf{R}_{t+1}^{b_k}\left(\boldsymbol{a}_{t+1} - \boldsymbol{b}_{a_t}\right)\times\left[\mathbf{I} - \delta t(\frac{\boldsymbol{\omega}_t+\boldsymbol{\omega}_{t+1}}{2} - \boldsymbol{b}_{g_t})\times\right]\right\}\delta\boldsymbol{\theta}_t^{b_k} \\
\quad +\frac{\delta t^2}{2}\mathbf{R}_{t+1}^{b_k}(\boldsymbol{a}_{t+1} - \boldsymbol{b}_{a_t})\times\delta\boldsymbol{b}_{g_t} - \frac{1}{2}\left(\mathbf{R}_t^{b_k} + \mathbf{R}_{t+1}^{b_k}\right)\delta\boldsymbol{b}_{a_t} + \frac{\delta t^2}{4}\mathbf{R}_{t+1}^{b_k}(\boldsymbol{a}_{t+1} - \boldsymbol{b}_{a_t})\times\boldsymbol{n}_{\omega_t} \\
\quad +\frac{\delta t^2}{4}\mathbf{R}_{t+1}^{b_k}(\boldsymbol{a}_{t+1} - \boldsymbol{b}_{a_t})\times\boldsymbol{n}_{\omega_{t+1}} - \frac{\delta t}{2}\mathbf{R}_t^{b_k}\boldsymbol{n}_{a_t} - \frac{\delta t}{2}\mathbf{R}_{t+1}^{b_k}\boldsymbol{n}_{a_{t+1}} + \delta\boldsymbol{\beta}_t^{b_k}
\end{cases}
\tag{21}
$$

$$
\begin{cases}
\dot{\delta\boldsymbol{\alpha}}_t^{b_k} = \frac{1}{2}\left(\delta\boldsymbol{\beta}_t^{b_k} + \delta\boldsymbol{\beta}_{t+1}^{b_k}\right) \\
\delta\boldsymbol{\alpha}_{t+1}^{b_k} = \delta\boldsymbol{\alpha}_t^{b_k} + \frac{1}{2}\left(\delta\boldsymbol{\beta}_t^{b_k} + \delta\boldsymbol{\beta}_{t+1}^{b_k}\right)\delta t
\end{cases}
$$

We can rewrite it in matrix form as follows (This is mostly according with the code. See "midPointIntegration" in "integration_base.h" in [6].),

$$
\begin{bmatrix} \delta\boldsymbol{\alpha}_{k+1} \\ \delta\boldsymbol{\theta}_{k+1} \\ \delta\boldsymbol{\beta}_{k+1} \\ \delta\boldsymbol{b}_{ak+1} \\ \delta\boldsymbol{b}_{gk+1} \end{bmatrix}
=
\begin{bmatrix} \mathbf{I} & \mathbf{F}_{01} & \delta t & \mathbf{F}_{03} & \mathbf{F}_{04} \\ \mathbf{0} & \mathbf{F}_{11} & \mathbf{0} & \mathbf{0} & -\delta t \\ \mathbf{0} & \mathbf{F}_{21} & \mathbf{I} & \mathbf{F}_{23} & \mathbf{F}_{24} \\ \mathbf{0} & \mathbf{0} & \mathbf{0} & \mathbf{I} & \mathbf{0} \\ \mathbf{0} & \mathbf{0} & \mathbf{0} & \mathbf{0} & \mathbf{I} \end{bmatrix}
\begin{bmatrix} \delta\boldsymbol{\alpha}_k \\ \delta\boldsymbol{\theta}_k \\ \delta\boldsymbol{\beta}_k \\ \delta\boldsymbol{b}_{ak} \\ \delta\boldsymbol{b}_{gk} \end{bmatrix}
+
\begin{bmatrix} \mathbf{G}_{00} & \mathbf{G}_{01} & \mathbf{G}_{02} & \mathbf{G}_{03} & \mathbf{0} & \mathbf{0} \\ \mathbf{0} & -\delta t/2 & \mathbf{0} & -\delta t/2 & \mathbf{0} & \mathbf{0} \\ \frac{-\mathbf{R}_k\delta t}{2} & \mathbf{G}_{21} & \frac{-\mathbf{R}_{k+1}\delta t}{2} & \mathbf{G}_{23} & \mathbf{0} & \mathbf{0} \\ \mathbf{0} & \mathbf{0} & \mathbf{0} & \mathbf{0} & \delta t & \mathbf{0} \\ \mathbf{0} & \mathbf{0} & \mathbf{0} & \mathbf{0} & \mathbf{0} & \delta t \end{bmatrix}
\begin{bmatrix} \boldsymbol{n}_{a_k} \\ \boldsymbol{n}_{g_k} \\ \boldsymbol{n}_{a_{k+1}} \\ \boldsymbol{n}_{g_{k+1}} \\ \boldsymbol{n}_{b_a} \\ \boldsymbol{n}_{b_g} \end{bmatrix}
\tag{22}
$$

where,

$$
\begin{aligned}
\mathbf{F}_{01} &= -\frac{\delta t^2}{4}\mathbf{R}_k\left(\hat{\boldsymbol{a}}_k-\boldsymbol{b}_{ak}\right)\times-\frac{\delta t^2}{4}\mathbf{R}_{k+1}\left(\hat{\boldsymbol{a}}_{k+1}-\boldsymbol{b}_{ak}\right)\times\left[\mathbf{I}-(\frac{\hat{\boldsymbol{\omega}}_k+\hat{\boldsymbol{\omega}}_{k+1}}{2}-\boldsymbol{b}_{\omega k})\times\delta t\right]\\
\mathbf{F}_{03} &= -\frac{\delta t^2}{4}\left(\mathbf{R}_k+\mathbf{R}_{k+1}\right)\\
\mathbf{F}_{04} &= \frac{\delta t^3}{4}\mathbf{R}_{k+1}\left(\hat{\boldsymbol{a}}_{k+1}-\boldsymbol{b}_{ak}\right)\times
\end{aligned}
\tag{23}
$$

$$
\mathbf{F}_{11}=\mathbf{I}-(\frac{\hat{\boldsymbol{\omega}}_k+\hat{\boldsymbol{\omega}}_{k+1}}{2}-\boldsymbol{b}_{gk})\times\delta t
\tag{24}
$$

$$
\begin{aligned}
\mathbf{F}_{21} &= -\frac{\delta t}{2}\mathbf{R}_k\left(\hat{\boldsymbol{a}}_k-\boldsymbol{b}_{ak}\right)\times-\frac{\delta t}{2}\mathbf{R}_{k+1}\left(\hat{\boldsymbol{a}}_{k+1}-\boldsymbol{b}_{ak}\right)\times\left[\mathbf{I}-\delta t(\frac{\hat{\boldsymbol{\omega}}_k+\hat{\boldsymbol{\omega}}_{k+1}}{2}-\boldsymbol{b}_{gk})\times\right]\\
\mathbf{F}_{23} &= -\frac{\delta t}{2}\left(\mathbf{R}_k+\mathbf{R}_{k+1}\right)\\
\mathbf{F}_{24} &= \frac{\delta t^2}{2}\mathbf{R}_{k+1}\left(\hat{\boldsymbol{a}}_{k+1}-\boldsymbol{b}_{ak}\right)\times
\end{aligned}
\tag{25}
$$

$$
\begin{aligned}
\mathbf{G}_{00} &= -\frac{\delta t^2}{4}\mathbf{R}_k\\
\mathbf{G}_{01} &= \mathbf{G}_{03}=\frac{\delta t^3}{8}\mathbf{R}_{k+1}\left(\hat{\boldsymbol{a}}_{k+1}-\boldsymbol{b}_{ak}\right)\times\\
\mathbf{G}_{02} &= -\frac{\delta t^2}{4}\mathbf{R}_{k+1}
\end{aligned}
\tag{26}
$$

$$
\mathbf{G}_{21}=\mathbf{G}_{23}=\frac{\delta t^2}{4}\mathbf{R}_{k+1}\left(\hat{\boldsymbol{a}}_{k+1}-\boldsymbol{b}_{ak}\right)\times
\tag{27}
$$

## III. INITIALIZATION

The initialization of the VINS-Mono can be divided into four main steps, as shown in Fig. 3.

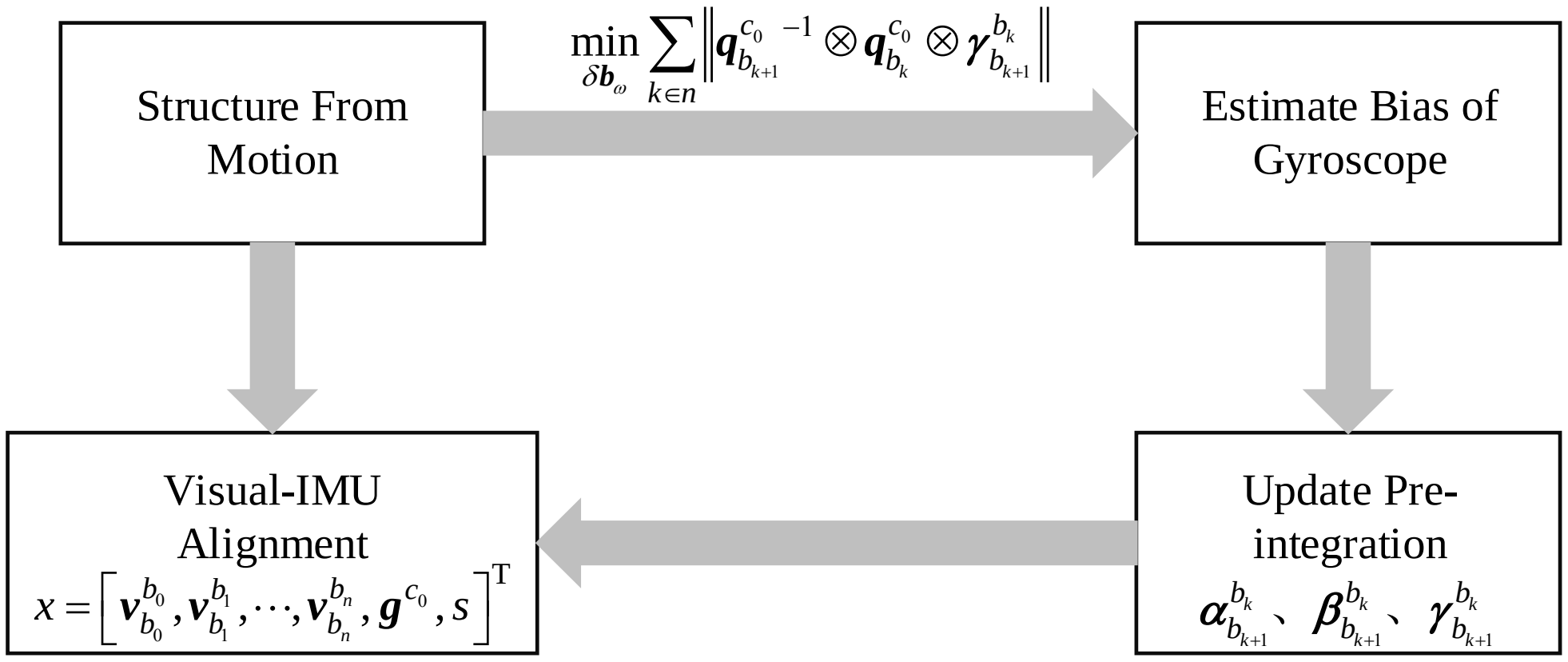

Fig. 3. Initialization procedure

It starts with a vision-only structure from motion (SFM) in the sliding window. Firstly, we recover the relative pose (up-to-scale) between two image frames by Five-point method when we can find a previous frame which has more than thirty tracked features, and the average parallax is more than twenty pixels between it and the latest frame. Then, we set this frame (not always the first frame in the code) as the reference frame ( $c_0$ ) for the moment. Subsequently, we triangulate all the features between the two frames. We perform EPnP [11] to estimate poses of other frames in the whole window based on the 3D features. At last, a bundle adjustment is performed to minimize the total reprojection error to optimize all poses in the window.

We can transform the poses from camera center to IMU center according to the extrinsic parameters $\begin{pmatrix} \boldsymbol{p}_c^b & \boldsymbol{q}_c^b \end{pmatrix}$

$$\begin{aligned} \boldsymbol{q}_{b_k}^{c_0} &= \boldsymbol{q}_{c_k}^{c_0} \otimes \left(\boldsymbol{q}_c^b\right)^{-1} \\ s\bar{\boldsymbol{p}}_{b_k}^{c_0} &= s\bar{\boldsymbol{p}}_{c_k}^{c_0} - \mathbf{R}_{b_k}^{c_0} \boldsymbol{p}_c^b \end{aligned} \tag{28}$$

where $\bar{\boldsymbol{p}}$ denotes the vector $\boldsymbol{p}$ up to the scale $s$ .

Changing the reference frame from the world frame $w$ to the $c_0$ frame in the SFM, we can rewrite the IMU preintegration in (5) as follow:

$$\begin{aligned} \hat{\boldsymbol{\alpha}}_{b_{k+1}}^{b_k} &= \mathbf{R}_{c_0}^{b_k} \left( s\left( \bar{\boldsymbol{p}}_{b_{k+1}}^{c_0} - \bar{\boldsymbol{p}}_{b_k}^{c_0} \right) + \frac{1}{2} \boldsymbol{g}^{c_0} \delta t^2 - \mathbf{R}_{b_k}^{c_0} \boldsymbol{v}_{b_k}^{b_k} \delta t \right) \\ \hat{\boldsymbol{\beta}}_{b_{k+1}}^{b_k} &= \mathbf{R}_{c_0}^{b_k} \left( \mathbf{R}_{b_{k+1}}^{c_0} \boldsymbol{v}_{b_{k+1}}^{b_{k+1}} + \boldsymbol{g}^{c_0} \delta t - \mathbf{R}_{b_k}^{c_0} \boldsymbol{v}_{b_k}^{b_k} \right) \end{aligned} \tag{29}$$

From (28) and (29) we can get a system of linear equations, which can be solved easily by matrix decomposition.

$$\begin{aligned} \hat{\boldsymbol{\alpha}}_{b_{k+1}}^{b_k} &= \mathbf{R}_{c_0}^{b_k} \left( s\left( \bar{\boldsymbol{p}}_{b_{k+1}}^{c_0} - \bar{\boldsymbol{p}}_{b_k}^{c_0} \right) + \frac{1}{2} \boldsymbol{g}^{c_0} \delta t^2 - \mathbf{R}_{b_k}^{c_0} \boldsymbol{v}_{b_k}^{b_k} \delta t \right) \\ &= \mathbf{R}_{c_0}^{b_k} \left( s\bar{\boldsymbol{p}}_{c_{k+1}}^{c_0} - \mathbf{R}_{b_{k+1}}^{c_0} \boldsymbol{p}_c^b - (s\bar{\boldsymbol{p}}_{c_k}^{c_0} - \mathbf{R}_{b_k}^{c_0} \boldsymbol{p}_c^b) + \frac{1}{2} \boldsymbol{g}^{c_0} \delta t^2 - \mathbf{R}_{b_k}^{c_0} \boldsymbol{v}_{b_k}^{b_k} \delta t \right) \\ &= \mathbf{R}_{c_0}^{b_k} \left( \bar{\boldsymbol{p}}_{c_{k+1}}^{c_0} - \bar{\boldsymbol{p}}_{c_k}^{c_0} \right) s + \frac{1}{2} \mathbf{R}_{c_0}^{b_k} \delta t^2 \boldsymbol{g}^{c_0} - \boldsymbol{v}_{b_k}^{b_k} \delta t + \boldsymbol{p}_c^b - \mathbf{R}_{c_0}^{b_k} \mathbf{R}_{b_{k+1}}^{c_0} \boldsymbol{p}_c^b \end{aligned} \tag{30}$$

$$
\begin{bmatrix} \hat{\boldsymbol{\alpha}}_{b_{k+1}}^{b_k} - \boldsymbol{p}_c^b + \mathbf{R}_{c_0}^{b_k}\mathbf{R}_{b_{k+1}}^{c_0}\boldsymbol{p}_c^b \\ \hat{\boldsymbol{\beta}}_{b_{k+1}}^{b_k} \end{bmatrix} =
$$

$$
\begin{bmatrix} -\delta t\mathbf{I} & 0 & \frac{1}{2}\mathbf{R}_{c_0}^{b_k}\delta t^2 & \mathbf{R}_{c_0}^{b_k}\left(\bar{\boldsymbol{p}}_{c_{k+1}}^{c_0} - \bar{\boldsymbol{p}}_{c_k}^{c_0}\right) \\ -\mathbf{I} & \mathbf{R}_{c_0}^{b_k}\mathbf{R}_{b_{k+1}}^{c_0} & \delta t\mathbf{R}_{c_0}^{b_k} & 0 \end{bmatrix} \begin{bmatrix} \boldsymbol{v}_{b_k}^{b_k} \\ \boldsymbol{v}_{b_{k+1}}^{b_{k+1}} \\ \boldsymbol{g}^{c_0} \\ s \end{bmatrix} \tag{31}
$$

The state we estimate in this step is

$$
\boldsymbol{\chi} = \left[\boldsymbol{v}_{b_0}^{b_0},\ \boldsymbol{v}_{b_1}^{b_1}, \cdots, \boldsymbol{v}_{b_n}^{b_n},\ \boldsymbol{g}^{c_0}, s\right] \tag{32}
$$

Moreover, the estimated gravity vector can be refined by the known gravity magnitude. Generally, a vector in a three dimensional space is constraint by its magnitude and projection on three axes. If we know any three values of them, the other one can be uniquely determined. Since the magnitude of gravity is known, the degrees of freedom of gravity is two. We can parameterize the gravity with two variables on its tangent space as follow

$$
\boldsymbol{g}^{c_0} = \|\boldsymbol{g}\| \frac{\boldsymbol{g}^{c_0}}{\left\|\boldsymbol{g}^{c_0}\right\|} + \begin{bmatrix}\boldsymbol{b}_1 & \boldsymbol{b}_2\end{bmatrix} \begin{bmatrix} w_1 \\ w_2 \end{bmatrix} \tag{33}
$$

where $\boldsymbol{g}^{c_0}$、$\boldsymbol{b}_1$、$\boldsymbol{b}_2$ are orthogonal; $\boldsymbol{b}_1$、$\boldsymbol{b}_2$ is generated by vector cross product. We substitute $\boldsymbol{g}^{c_0}$ in Eq. (31) by Eq. (33), and solve $w_1$ and $w_2$ recursively until converge (when $w_1$ and $w_2$ are close to zero).

After refining gravity, we can recover the transformation from the $c_0$ frame to the world frame by rotating the gravity vector. We set the origin of the world frame at that of the $c_0$ frame. However, the heading angle can be any value because the number of unknown terms in the rotation matrix is nine (which is more than the knowns), and we cannot determine the absolute heading using a camera and a MEMS IMU. So we can rotate the world frame to ensure the heading angle of the first frame is zero.

Now, the initialization process is completed. We have recovered the scale of visual odometry by aligning the IMU measurements with visual-based SFM. We have established the world frame whose z axis is parallel with the gravity vector. Additionally, we have estimated the real-scaled velocity in the world frame.

## IV. Tightly-coupled Nonlinear Optimization

### *4.1 Cost Function*

In the back-end optimization, the inverse depth of features (Why inverse depth? Because it has better numerical stability which is convenient to solve [12]), the pose and velocity of every frame, the IMU bias (gyroscope and accelerator) and the extrinsic parameters are optimized together.

The state estimation can be thought as a Maximum A Posteriori (MAP) problem [13]. The estimation of the state is equal to calculating the conditional probability distribution of state quantities under conditions of known observations:

$$P\left(\boldsymbol{x}|\boldsymbol{z}\right) \tag{34}$$

According to the Bayesian rule, we have

$$P\left(\boldsymbol{x}|\boldsymbol{z}\right)=\frac{P\left(\boldsymbol{z}|\boldsymbol{x}\right)P\left(\boldsymbol{x}\right)}{P\left(\boldsymbol{z}\right)}\propto P\left(\boldsymbol{z}|\boldsymbol{x}\right)P\left(\boldsymbol{x}\right) \tag{35}$$

where $P\left(\boldsymbol{x}|\boldsymbol{z}\right)$ is the posterior probability; $P\left(\boldsymbol{x}\right)$ is the priori probability; $P\left(\boldsymbol{z}|\boldsymbol{x}\right)$ is the likelihood probability. It is difficult to calculate the posterior probability distribution directly. However, it is feasible to find an optimal state estimation which maximizes the posterior probability. In most cases, we do not have any priori information about the state of the system. So, we have

$$\boldsymbol{\chi}^*=\arg\max P\left(\boldsymbol{x}|\boldsymbol{z}\right)\propto \arg\max P\left(\boldsymbol{z}|\boldsymbol{x}\right)P\left(\boldsymbol{x}\right)\propto \arg\max P\left(\boldsymbol{z}|\boldsymbol{x}\right) \tag{36}$$

The problem has evolved to finding a state of the system that can produce the observations most probably.

We assume the uncertainty of measurement is Gaussian distributed, namely, $\boldsymbol{z}\sim\boldsymbol{N}(\overline{\boldsymbol{z}},\boldsymbol{Q})$. Then, the negative log-likelihood of (36) can be written as

$$\boldsymbol{\chi}^*=\arg\max P\left(\boldsymbol{z}|\boldsymbol{x}\right)=\arg\min\sum\left\|\boldsymbol{z}-h(\boldsymbol{x})\right\|_{\boldsymbol{Q}} \tag{37}$$

$h(\bullet)$ is a function of the state. $\left\|\bullet\right\|_{\boldsymbol{Q}}$ denotes the Mahalanobis norm.

**NOTICE:** Considering an arbitrary dimension Gaussian distribution, $\boldsymbol{x}\sim\boldsymbol{N}\left(\boldsymbol{\mu},\boldsymbol{Q}\right)$, the probability density function of $\boldsymbol{x}$ can be written as

$$P(\boldsymbol{x})=\frac{1}{\sqrt{(2\pi)^N\det(\boldsymbol{Q})}}\exp(-\frac{1}{2}(\boldsymbol{x}-\boldsymbol{\mu})^T\boldsymbol{Q}^{-1}(\boldsymbol{x}-\boldsymbol{\mu})) \tag{38}$$

$$-\ln P(\boldsymbol{x})=\ln\sqrt{(2\pi)^N\det(\boldsymbol{Q})}+\frac{1}{2}(\boldsymbol{x}-\boldsymbol{\mu})^T\boldsymbol{Q}^{-1}(\boldsymbol{x}-\boldsymbol{\mu}) \tag{39}$$

Since the first part in the left side does not contain $\boldsymbol{x}$, so to maximize $P(\boldsymbol{x})$ become to minimize the second part in the right side of Eq. (39).

In the VINS-Mono, the cost function can be written as

$$\min_{\chi}\left\{\left\|\boldsymbol{r}_p-\mathbf{J}_P\boldsymbol{\chi}\right\|_{\mathbf{P}_M}+\sum_{k\in\boldsymbol{B}}\left\|\boldsymbol{r}_{\boldsymbol{B}}\left(\hat{\boldsymbol{z}}_{b_{k+1}}^{b_k},\boldsymbol{\chi}\right)\right\|_{\mathbf{P}_B}+\sum_{(l,j)\in\boldsymbol{C}}\left\|\boldsymbol{r}_{\boldsymbol{C}}\left(\hat{\boldsymbol{z}}_l^{c_j},\boldsymbol{\chi}\right)\right\|_{\mathbf{P}_l^{c_j}}\right\} \tag{40}$$

where $\left(\boldsymbol{r}_p, \ \mathbf{J}_p\right)$ is the priori information from marginalization; $\boldsymbol{r}_{\boldsymbol{B}}$ is the residuals of the IMU measurements; $\boldsymbol{B}$ is the set of all IMU measurements in the sliding window; $\boldsymbol{r}_{\boldsymbol{C}}$ is the residuals of the visual model; and $\boldsymbol{C}$ is the set of features which have been observed at least two times in the sliding window.

The state estimation is converted to a nonlinear least square problem, and it can be solved by Gaussian-Newton or Levenberg-Marquardt approach.

We take the Gaussian-Newton approach for example to solve a general optimization problem,

$$\chi = \underset{\boldsymbol{x}}{\operatorname{argmin}} \left\| f\left(\boldsymbol{x}\right) \right\|_{\mathbf{P}} \tag{41}$$

where $\mathbf{P}$ is the covariance matrix of the residuals; $\|\bullet\|$ denotes the two-norm. We take into account the first-order Taylor expansion of the cost function $f\left(x\right)$, then the problem become calculating the increment $\delta\boldsymbol{x}$.

$$\begin{aligned} &\delta\boldsymbol{x}^* = \underset{\delta\boldsymbol{x}}{\operatorname{argmin}} \left\| f\left(\boldsymbol{x}\right) + \mathbf{J}(x)\delta\boldsymbol{x} \right\|_{\mathbf{P}} \\ &\left\| f\left(\boldsymbol{x}\right) + \mathbf{J}(x)\delta\boldsymbol{x} \right\|_{\mathbf{P}} = \left( f\left(\boldsymbol{x}\right) + \mathbf{J}\left(\boldsymbol{x}\right)\delta\boldsymbol{x}\right)^{\mathrm{T}} \mathbf{P}^{-1} \left( f\left(\boldsymbol{x}\right) + \mathbf{J}\left(\boldsymbol{x}\right)\delta\boldsymbol{x}\right) \\ &= f\left(\boldsymbol{x}\right)^{\mathrm{T}} \mathbf{P}^{-1} f\left(\boldsymbol{x}\right) + 2\mathbf{J}\left(\boldsymbol{x}\right)^{\mathrm{T}} \mathbf{P}^{-1} f\left(\boldsymbol{x}\right)\delta\boldsymbol{x} + \delta\boldsymbol{x}^{\mathrm{T}} \mathbf{J}\left(\boldsymbol{x}\right)^{\mathrm{T}} \mathbf{P}^{-1} \mathbf{J}\left(\boldsymbol{x}\right)\delta\boldsymbol{x} \end{aligned} \tag{42}$$

Calculating the derivative of the function in (42) and let it equal to zero, we have

$$\begin{gathered} \mathbf{J}\left(\boldsymbol{x}\right)^{\mathrm{T}} \mathbf{P}^{-1} \mathbf{J}\left(\boldsymbol{x}\right)\delta\boldsymbol{x} = -\mathbf{J}\left(\boldsymbol{x}\right)^{\mathrm{T}} \mathbf{P}^{-1} f\left(\boldsymbol{x}\right) \\ \mathbf{H}\delta\boldsymbol{x} = \boldsymbol{b} \end{gathered} \tag{43}$$

This can be solved by matrix decomposition, e.g. SVD.

*4.2 IMU Model*

As per Eq. (5), we can write the residuals of IMU measurements as following,

$$\boldsymbol{r}_{\boldsymbol{B}}\left(\hat{\boldsymbol{z}}_{b_{k+1}}^{b_k}, \chi\right) = \begin{bmatrix} \delta\boldsymbol{\alpha}_{b_{k+1}}^{b_k} \\ \delta\boldsymbol{\theta}_{b_{k+1}}^{b_k} \\ \delta\boldsymbol{\beta}_{b_{k+1}}^{b_k} \\ \delta\boldsymbol{b}_a \\ \delta\boldsymbol{b}_g \end{bmatrix} = \begin{bmatrix} \mathbf{R}_w^{b_k}(\boldsymbol{p}_{b_{k+1}}^w - \boldsymbol{p}_{b_k}^w - \boldsymbol{v}_{b_k}^w \Delta t_k + \frac{1}{2}\boldsymbol{g}^w \Delta t_k{}^2) - \hat{\boldsymbol{\alpha}}_{b_{k+1}}^{b_k} \\ 2\left[\left(\hat{\boldsymbol{\gamma}}_{b_{k+1}}^{b_k}\right)^{-1} \otimes (\boldsymbol{q}_{b_k}^w)^{-1} \otimes \boldsymbol{q}_{b_{k+1}}^w\right]_{xyz} \\ \mathbf{R}_w^{b_k}(\boldsymbol{v}_{b_{k+1}}^w - \boldsymbol{v}_{b_k}^w + \boldsymbol{g}^w \Delta t_k) - \hat{\boldsymbol{\beta}}_{b_{k+1}}^{b_k} \\ \boldsymbol{b}_{ak+1} - \boldsymbol{b}_{ak} \\ \boldsymbol{b}_{gk+1} - \boldsymbol{b}_{gk} \end{bmatrix} \tag{44}$$

For IMU residual model, the optimization variables are

$$\left[\boldsymbol{p}_{b_k}^w \quad \boldsymbol{q}_{b_k}^w\right]\left[\boldsymbol{v}_{b_k}^w \quad \boldsymbol{b}_{ak} \quad \boldsymbol{b}_{gk}\right]\left[\boldsymbol{p}_{b_{k+1}}^w \quad \boldsymbol{q}_{b_{k+1}}^w\right]\left[\boldsymbol{v}_{b_{k+1}}^w \quad \boldsymbol{b}_{ak+1} \quad \boldsymbol{b}_{gk+1}\right] \tag{45}$$

It is consistent with the code implementation. The dimensions of the state vectors above are 7, 9, 7, and 9, respectively.

**NOTICE:** In the state vector, we use the quaternion to represent the attitude of the vehicle. However, it is not convenient to calculate the Jacobian w.r.t the attitude in quaternion. In addition, we know that the degrees of attitude is three while there are four variables in the quaternion which can overparameterize the problem. So we use Lie-algebra $so(3)$ (namely, rotation vector) to perform perturbation analysis and iterative optimization when considering the attitude error. We calculate the increment of attitude in $so(3)$, then we update the states by the quaternion of the attitude increment. This is implemented by instantiating "LocalParameterization" in ceres solver [14]. Therefore, the dimensions of the Jacobian matrix in the IMU residual model should be <15×7, 15×9, 15×7, 15×9>. Therefore the last row of $\mathbf{J}[0]$ and $\mathbf{J}[2]$ is zero. For a simple expression, we only write the left six rows of them (see Eq. (46) and (48)). The vison model is written as the same way.

The explanation of the attitude perturbation analysis is shown in Appendix A. We perform perturbation analysis for Eq. (44), and the Jacobian matrix can be written as follows. The details can be found in Appendix B.

$$\mathbf{J}[0]_{15\times 6}=\frac{\partial \boldsymbol{r}_{\boldsymbol{B}}\left(\hat{\boldsymbol{z}}_{b_{k+1}}^{b_k},\boldsymbol{\chi}\right)}{\partial\left[\boldsymbol{p}_{b_k}^{w}\quad \boldsymbol{q}_{b_k}^{w}\right]}=\begin{bmatrix} -\mathbf{R}_w^{b_k} & \left[\mathbf{R}_w^{b_k}(\boldsymbol{p}_{b_{k+1}}^{w}-\boldsymbol{p}_{b_k}^{w}-\boldsymbol{v}_{b_k}^{w}\Delta t_k+\frac{1}{2}\boldsymbol{g}^{w}\Delta t_k^{\ 2})\right]\times \\ \mathbf{0} & -R\left[\boldsymbol{\gamma}_{b_{k+1}}^{b_k}\right]L\left[\left(\boldsymbol{q}_{b_{k+1}}^{w}\right)^{-1}\otimes \boldsymbol{q}_{w}^{b_k}\right]_{3\times 3} \\ \mathbf{0} & \left[\mathbf{R}_w^{b_k}(\boldsymbol{p}_{b_{k+1}}^{w}-\boldsymbol{p}_{b_k}^{w}+\boldsymbol{g}^{w}\Delta t_k)\right]\times \\ \mathbf{0} & \mathbf{0} \\ \mathbf{0} & \mathbf{0} \end{bmatrix} \tag{46}$$

$$\mathbf{J}[1]_{15\times 9}=\frac{\partial \boldsymbol{r}_{\boldsymbol{B}}\left(\hat{\boldsymbol{z}}_{b_{k+1}}^{b_k},\boldsymbol{\chi}\right)}{\partial\left[\boldsymbol{v}_{b_k}^{w}\quad \boldsymbol{b}_{ak}\quad \boldsymbol{b}_{gk}\right]}=\begin{bmatrix} -\mathbf{R}_w^{b_k}\Delta t_k & -\mathbf{J}_{b_a}^{\alpha} & -\mathbf{J}_{b_g}^{\alpha} \\ \mathbf{0} & \mathbf{0} & -R\left[\left(\hat{\boldsymbol{\gamma}}_{b_{k+1}}^{b_k}\right)^{-1}\otimes(\boldsymbol{q}_{b_k}^{w})^{-1}\otimes \boldsymbol{q}_{b_{k+1}}^{w}\right]_{3\times 3}\mathbf{J}_{b_g}^{\gamma} \\ -\mathbf{R}_w^{b_k} & -\mathbf{J}_{b_a}^{\beta} & -\mathbf{J}_{b_g}^{\beta} \\ \mathbf{0} & -\mathbf{I} & \mathbf{0} \\ \mathbf{0} & \mathbf{0} & -\mathbf{I} \end{bmatrix}$$

(47)

$$\mathbf{J}[2]_{15\times 6} = \frac{\partial \boldsymbol{r_B}\left(\hat{\boldsymbol{z}}_{b_{k+1}}^{b_k}, \chi\right)}{\partial\left[\boldsymbol{p}_{b_{k+1}}^{w} \quad \boldsymbol{q}_{b_{k+1}}^{w}\right]} = \begin{bmatrix} \mathbf{R}_w^{b_k} & \mathbf{0} \\ \mathbf{0} & L\left[\left(\hat{\gamma}_{b_{k+1}}^{b_k}\right)^{-1} \otimes (\boldsymbol{q}_{b_k}^{w})^{-1} \otimes \boldsymbol{q}_{b_{k+1}}^{w}\right] \\ \mathbf{0} & \mathbf{0} \\ \mathbf{0} & \mathbf{0} \\ \mathbf{0} & \mathbf{0} \end{bmatrix} \tag{48}$$

$$\mathbf{J}[3]_{15\times 9} = \frac{\partial \boldsymbol{r_B}\left(\hat{\boldsymbol{z}}_{b_{k+1}}^{b_k}, \chi\right)}{\partial\left[\boldsymbol{v}_{b_{k+1}}^{w} \quad \boldsymbol{b}_{ak+1} \quad \boldsymbol{b}_{gk+1}\right]} = \begin{bmatrix} \mathbf{0} & \mathbf{0} & \mathbf{0} \\ \mathbf{0} & \mathbf{0} & \mathbf{0} \\ \mathbf{R}_w^{b_k} & \mathbf{0} & \mathbf{0} \\ \mathbf{0} & \mathbf{I} & \mathbf{0} \\ \mathbf{0} & \mathbf{0} & \mathbf{I} \end{bmatrix} \tag{49}$$

*4.3 Vision Model*

Considering the $l^{th}$ feature which is firstly observed in $i^{th}$ image, the residual for the observation in the $j^{th}$ image can be written as

$$\hat{\boldsymbol{P}}_l^{c_j} = \boldsymbol{\pi}_c^{-1}\begin{bmatrix} \hat{u}_l^{c_j} \\ \hat{v}_l^{c_j} \end{bmatrix}$$

$$\tilde{\boldsymbol{P}}_l^{c_j} = \mathbf{R}_b^c\left\{\mathbf{R}_w^{b_j}\left[\mathbf{R}_{b_i}^w\left(\mathbf{R}_c^b \frac{1}{\lambda_l}\boldsymbol{\pi}_c^{-1}\begin{bmatrix} \hat{u}_l^{c_i} \\ \hat{v}_l^{c_i} \end{bmatrix} + \boldsymbol{P}_c^b\right) + \boldsymbol{P}_{b_i}^w\right] + \boldsymbol{P}_w^{b_j}\right\} + \boldsymbol{P}_b^c$$

$$= \mathbf{R}_b^c\left\{\mathbf{R}_w^{b_j}\left[\mathbf{R}_{b_i}^w\left(\mathbf{R}_c^b \frac{1}{\lambda_l}\boldsymbol{\pi}_c^{-1}\begin{bmatrix} \hat{u}_l^{c_i} \\ \hat{v}_l^{c_i} \end{bmatrix} + \boldsymbol{P}_c^b\right) + \boldsymbol{P}_{b_i}^w - \boldsymbol{P}_{b_j}^w\right] - \boldsymbol{P}_c^b\right\}$$

$$\boldsymbol{r_C}\left(\hat{\boldsymbol{z}}_l^{c_j}, \chi\right) = [\boldsymbol{b}_1, \boldsymbol{b}_2]^{\mathrm{T}} \bullet \left(\frac{\tilde{\boldsymbol{P}}_l^{c_j}}{\left\|\tilde{\boldsymbol{P}}_l^{c_j}\right\|} - \hat{\boldsymbol{P}}_l^{c_j}\right)$$

*or*

$$\boldsymbol{r_C}\left(\hat{\boldsymbol{z}}_l^{c_j}, \chi\right) = \left(\frac{\tilde{\boldsymbol{P}}_l^{c_j}}{\tilde{\boldsymbol{P}}_l^{c_j}[2]} - \hat{\boldsymbol{P}}_l^{c_j}\right)_{head\, 2\times 1} \tag{50}$$

let

$$\bar{\boldsymbol{P}}_l^{c_i} = \boldsymbol{\pi}_c^{-1}\begin{bmatrix} \hat{u}_l^{c_i} \\ \hat{v}_l^{c_i} \end{bmatrix} \tag{51}$$

where $\boldsymbol{\pi}_c^{-1}$ is the back projection model which outputs the correspondence **normalized vector** in 3D space(see "projection_factor.cpp" in the code); $\left[\hat{u}_l^{c_i}, \hat{v}_l^{c_i}\right]$ is the first observation of the $l^{th}$ feature in the $i^{th}$ image; $\left[\hat{u}_l^{c_j}, \hat{v}_l^{c_j}\right]$ is the observation of the $l^{th}$ feature in the $j^{th}$ image. The authors project the residual vector onto the tangent plane. $[\boldsymbol{b}_1, \boldsymbol{b}_2]$ are two arbitrarily selected orthogonal bases which span the tangent plane of $\hat{\boldsymbol{P}}_l^{c_j}$.

**NOTICE:** The second equation of (50) can be derived by the following equations.

$$\begin{aligned} &\mathbf{R}_b^w \boldsymbol{P}^b + \boldsymbol{P}_b^w = \boldsymbol{P}^w \\ &\boldsymbol{P}_w^b = -\mathbf{R}_w^b \boldsymbol{P}_b^w \end{aligned} \tag{52}$$

For Vision model, the optimization variables are

$$\begin{bmatrix} \boldsymbol{p}_{b_i}^w & \boldsymbol{q}_{b_i}^w \end{bmatrix} \begin{bmatrix} \boldsymbol{p}_{b_j}^w & \boldsymbol{q}_{b_j}^w \end{bmatrix} \begin{bmatrix} \boldsymbol{p}_c^b & \boldsymbol{q}_c^b \end{bmatrix}, \lambda_l \tag{53}$$

The Jacobian can be calculated by the chain rule,

$$\frac{\partial \mathbf{r}_C}{\partial \mathbf{x}} = \frac{\partial \mathbf{r}_C}{\partial \tilde{\boldsymbol{P}}_l^{c_j}} \bullet \frac{\partial \tilde{\boldsymbol{P}}_l^{c_j}}{\partial \mathbf{x}} \tag{54}$$

where

$$\frac{\partial \mathbf{r}_C}{\partial \tilde{\boldsymbol{P}}_l^{c_j}} = \begin{bmatrix} \dfrac{1}{z_l^{c_j}} & 0 & -\dfrac{x_l^{c_j}}{\left(z_l^{c_j}\right)^2} \\ 0 & \dfrac{1}{z_l^{c_j}} & -\dfrac{y_l^{c_j}}{\left(z_l^{c_j}\right)^2} \end{bmatrix} \tag{55}$$

For the tangent space form, we have

$$\frac{\partial \boldsymbol{r}_C}{\partial \tilde{\boldsymbol{P}}_l^{c_j}} = \left[\boldsymbol{b}_1, \boldsymbol{b}_2\right]^{\mathrm{T}} \bullet \frac{\partial \left( \frac{\tilde{\boldsymbol{P}}_l^{c_j}}{\left\| \tilde{\boldsymbol{P}}_l^{c_j} \right\|} - \hat{\boldsymbol{P}}_l^{c_j} \right)}{\partial \tilde{\boldsymbol{P}}_l^{c_j}}$$

$$= \frac{\partial \frac{\tilde{\boldsymbol{P}}_l^{c_j}}{\left\| \tilde{\boldsymbol{P}}_l^{c_j} \right\|}}{\partial \tilde{\boldsymbol{P}}_l^{c_j}} = \frac{1}{\left\| \tilde{\boldsymbol{P}}_l^{c_j} \right\|} \mathbf{I} - \frac{\tilde{\boldsymbol{P}}_l^{c_j} \frac{\partial \left\| \tilde{\boldsymbol{P}}_l^{c_j} \right\|}{\partial \tilde{\boldsymbol{P}}_l^{c_j}}}{\left\| \tilde{\boldsymbol{P}}_l^{c_j} \right\|^2}$$

$$= \begin{bmatrix} \frac{1}{\left\| \tilde{\boldsymbol{P}}_l^{c_j} \right\|} - \frac{x^2}{\left\| \tilde{\boldsymbol{P}}_l^{c_j} \right\|^3} & -\frac{xy}{\left\| \tilde{\boldsymbol{P}}_l^{c_j} \right\|^3} & -\frac{xz}{\left\| \tilde{\boldsymbol{P}}_l^{c_j} \right\|^3} \\ -\frac{xy}{\left\| \tilde{\boldsymbol{P}}_l^{c_j} \right\|^3} & \frac{1}{\left\| \tilde{\boldsymbol{P}}_l^{c_j} \right\|} - \frac{y^2}{\left\| \tilde{\boldsymbol{P}}_l^{c_j} \right\|^3} & -\frac{yz}{\left\| \tilde{\boldsymbol{P}}_l^{c_j} \right\|^3} \\ -\frac{xz}{\left\| \tilde{\boldsymbol{P}}_l^{c_j} \right\|^3} & -\frac{yz}{\left\| \tilde{\boldsymbol{P}}_l^{c_j} \right\|^3} & \frac{1}{\left\| \tilde{\boldsymbol{P}}_l^{c_j} \right\|} - \frac{z^2}{\left\| \tilde{\boldsymbol{P}}_l^{c_j} \right\|^3} \end{bmatrix} \tag{56}$$

$$\left( let \quad \tilde{\boldsymbol{P}}_l^{c_j} = \left[ x \quad y \quad z \right]^{\mathrm{T}} \right)$$

For $\frac{\partial \tilde{\boldsymbol{P}}_l^{c_j}}{\partial \mathbf{x}}$, we have the following equations. Details can be found in Appendix C.

$$\mathbf{J}\left[0\right]_{3\times 6} = \frac{\partial \tilde{\boldsymbol{P}}_l^{c_j}}{\partial \left[ \boldsymbol{p}_{b_i}^w \quad \boldsymbol{q}_{b_i}^w \right]} = \left[ \mathbf{R}_b^c \mathbf{R}_w^{b_j} \quad -\mathbf{R}_b^c \mathbf{R}_w^{b_j} \mathbf{R}_{b_i}^w \left[ \left( \frac{1}{\lambda_l} \mathbf{R}_c^b \bar{\boldsymbol{P}}_l^{c_i} + \boldsymbol{P}_c^b \right) \times \right] \right] \tag{57}$$

$$\mathbf{J}\left[1\right]_{3\times 6} = \frac{\partial \tilde{\boldsymbol{P}}_l^{c_j}}{\partial \left[ \boldsymbol{p}_{b_j}^w \quad \boldsymbol{q}_{b_j}^w \right]} = \left[ -\mathbf{R}_b^c \mathbf{R}_w^{b_j} \quad \mathbf{R}_b^c \left\{ \left[ \mathbf{R}_w^{b_j} \left( \mathbf{R}_{b_i}^w \left( \mathbf{R}_c^b \frac{1}{\lambda_l} \bar{\boldsymbol{P}}_l^{c_i} \right) + \boldsymbol{P}_{b_i}^w - \boldsymbol{P}_{b_j}^w \right) \right] \times \right\} \right] \tag{58}$$

$$\mathbf{J}\left[2\right]_{3\times 6} = \frac{\partial \tilde{\boldsymbol{P}}_l^{c_j}}{\partial \left[ \boldsymbol{p}_c^b \quad \boldsymbol{q}_c^b \right]} =$$

$$\begin{bmatrix} \mathbf{R}_b^c \left( \mathbf{R}_w^{b_j} \mathbf{R}_{b_i}^w - \mathbf{I} \right) & \left[ \mathbf{R}_b^c \left( \mathbf{R}_w^{b_j} \left[ \mathbf{R}_{b_i}^w \boldsymbol{P}_c^b + \boldsymbol{P}_{b_i}^w - \boldsymbol{P}_{b_j}^w \right] - \boldsymbol{P}_c^b \right) \right] \times + \\ & \left( \mathbf{R}_b^c \mathbf{R}_w^{b_j} \mathbf{R}_{b_i}^w \mathbf{R}_c^b \frac{1}{\lambda_l} \bar{\boldsymbol{P}}_l^{c_i} \right) \times - \mathbf{R}_b^c \mathbf{R}_w^{b_j} \mathbf{R}_{b_i}^w \mathbf{R}_c^b \left( \frac{1}{\lambda_l} \bar{\boldsymbol{P}}_l^{c_i} \times \right) \end{bmatrix} \tag{59}$$

$$\mathbf{J}\left[3\right]_{3\times 1} = \frac{\partial \tilde{\boldsymbol{P}}_l^{c_j}}{\partial \lambda_l} = \left[ -\frac{1}{\lambda_l^2} \mathbf{R}_b^c \mathbf{R}_w^{b_j} \mathbf{R}_{b_i}^w \mathbf{R}_c^b \bar{\boldsymbol{P}}_l^{c_i} \right] \tag{60}$$

## V. MARGINALIZATION

Since the number of states increase along with the time, the computational complexity will increase quadratically accordingly. In order to reduce compute burden without loss of information, the marginalization procedure is performed to convert the previous measurements into a prior term. The set of states to be marginalized is denoted as $\chi_m$ and the set of remaining states is denoted as $\chi_r$ . According to (43), we rearrange the states' order and get the following equation

$$\begin{bmatrix} \mathbf{H}_{mm} & \mathbf{H}_{mr} \\ \mathbf{H}_{rm} & \mathbf{H}_{rr} \end{bmatrix} \begin{bmatrix} \delta \boldsymbol{x}_m \\ \delta \boldsymbol{x}_r \end{bmatrix} = \begin{bmatrix} \boldsymbol{b}_m \\ \boldsymbol{b}_r \end{bmatrix} \tag{61}$$

Then we perform the Schur complement to carry out the marginalization [15, 16] as follows

$$\begin{gathered} \begin{bmatrix} \mathbf{I} & \mathbf{0} \\ -\mathbf{H}_{rm}\mathbf{H}_{mm}^{-1} & \mathbf{I} \end{bmatrix} \begin{bmatrix} \mathbf{H}_{mm} & \mathbf{H}_{mr} \\ \mathbf{H}_{rm} & \mathbf{H}_{rr} \end{bmatrix} \begin{bmatrix} \delta \boldsymbol{x}_m \\ \delta \boldsymbol{x}_r \end{bmatrix} = \begin{bmatrix} \mathbf{I} & \mathbf{0} \\ -\mathbf{H}_{rm}\mathbf{H}_{mm}^{-1} & \mathbf{I} \end{bmatrix} \begin{bmatrix} \boldsymbol{b}_m \\ \boldsymbol{b}_r \end{bmatrix} \\ \begin{bmatrix} \mathbf{H}_{mm} & \mathbf{H}_{mr} \\ \mathbf{0} & \underbrace{-\mathbf{H}_{rm}\mathbf{H}_{mm}^{-1}\mathbf{H}_{mr} + \mathbf{H}_{rr}}_{\mathbf{H}_p} \end{bmatrix} \begin{bmatrix} \delta \boldsymbol{x}_m \\ \delta \boldsymbol{x}_r \end{bmatrix} = \begin{bmatrix} \boldsymbol{b}_m \\ \underbrace{-\mathbf{H}_{rm}\mathbf{H}_{mm}^{-1}\boldsymbol{b}_m + \boldsymbol{b}_r}_{\boldsymbol{b}_p} \end{bmatrix} \end{gathered} \tag{62}$$

$$\left(-\mathbf{H}_{rm}\mathbf{H}_{mm}^{-1}\mathbf{H}_{mr} + \mathbf{H}_{rr}\right)\delta \boldsymbol{x}_r = -\mathbf{H}_{rm}\mathbf{H}_{mm}^{-1}\boldsymbol{b}_m + \boldsymbol{b}_r \tag{63}$$

Note that we keep states from instant $m$ to instant $n$ in the sliding window. The states before $m$ are marginalized out and converted to a prior term. Therefore, the MAP problem can be written as that in [16],

$$\chi_{m:n}{}^* = \arg\min_{\chi_{m:n}} \sum_{t=m}^{n} \sum_{k \in \mathbf{S}} \left\{ \left\| \boldsymbol{z}_t^k - h_t^k(\chi_{m:n}) \right\|_{\boldsymbol{\Omega}_t^k}^2 + (\mathbf{H}_p \delta \boldsymbol{x}_r - \boldsymbol{b}_p) \right\} \tag{64}$$

Where $\mathbf{S}$ is the set of measurements.

## VI. GLOBAL OPTIMIZATION IN THE VINS-FUSION

A general optimization-based framework for global pose estimation is proposed in [17], which is an extension of [5]. Local estimation from VIO/VO is fused with global sensor data in a pose graph optimization manner. Within the optimization, the transformation from the local frame to the global frame is estimated, so the local state can be aligned into the global coordinate system. The illustration of the global pose graph structure is shown in Fig. 4. The state vector to be estimated in global optimization is shown in Eq. (65).

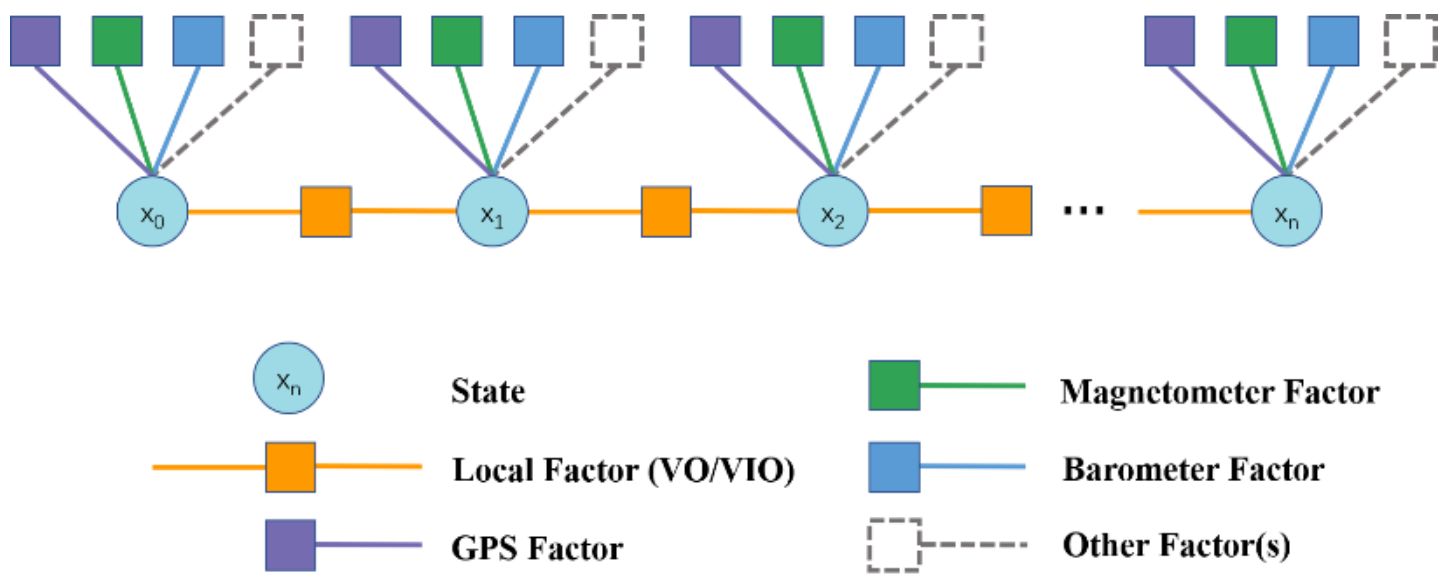

Fig. 4 Illustration of the global pose graph structure. (Tong Qin, 2018)

$$\chi = \left[ \boldsymbol{q}_i^G、\boldsymbol{P}_i^G、\cdots、\boldsymbol{q}_n^G、\boldsymbol{P}_n^G \right] \tag{65}$$

Every node contains the pose of the vehicle in the global frame (Here is the GPS frame), while the edge between two consecutive nodes is a local constraint from VIO/VO estimation. The local pose and global pose of one node can be written as a combination of rotation and translation,

$$\begin{aligned} \mathbf{T}_i^l &= \left[ \mathbf{R}_i^l \quad \boldsymbol{P}_i^l \right] \\ \mathbf{T}_i^G &= \left[ \mathbf{R}_i^G \quad \boldsymbol{P}_i^G \right] \end{aligned} \tag{66}$$

where $i$ is the $i^{th}$ node; $l$ is the local reference frame; $G$ is the global reference frame; $\mathbf{R}$ and $\mathbf{T}$ are the rotation and translation from body frame to reference frame, respectively. Given two consecutive nodes $i$ and $j$, the relative transformation can be derived from local pose and global pose, respectively. In the code, the authors set the first GPS point as the origin and subsequently set the ENU (East-North-Up) coordinate system as global frame.

$$\mathbf{T}_{j_l}^i = \begin{bmatrix} \mathbf{R}_i^l & \boldsymbol{P}_i^l \\ 0 & 1 \end{bmatrix}^{-1} \begin{bmatrix} \mathbf{R}_j^l & \boldsymbol{P}_j^l \\ 0 & 1 \end{bmatrix} = \begin{bmatrix} \mathbf{R}_{j_l}^i & \boldsymbol{P}_{ij_l}^i \\ 0 & 1 \end{bmatrix} \tag{67}$$

$$\mathbf{T}_{j_G}^i = \begin{bmatrix} \mathbf{R}_G^i \mathbf{R}_j^G & \mathbf{R}_G^i \left( \boldsymbol{P}_j^G - \boldsymbol{P}_i^G \right) \\ 0 & 1 \end{bmatrix} = \begin{bmatrix} \mathbf{R}_{j_G}^i & \boldsymbol{P}_{ij_G}^i \\ 0 & 1 \end{bmatrix} \tag{68}$$

Then, the residuals can be written as

$$\begin{bmatrix} \delta \boldsymbol{P}_{ij} \\ \delta \boldsymbol{\theta}_{ij} \end{bmatrix} = \begin{bmatrix} \boldsymbol{P}_{ij_G}^i - \boldsymbol{P}_{ij_l}^i \\ 2\left( \boldsymbol{q}_{j_l}^i \right)^{-1} \boldsymbol{q}_{j_G}^i \end{bmatrix} \tag{69}$$

For GPS position constraint, we have

$$\delta \boldsymbol{P}_i = \left[ \boldsymbol{P}_{i_estimation}^G - \boldsymbol{P}_{i_measurement}^G \right] \tag{70}$$

The standard deviation (std) of the global position error can be obtained from GPS positioning algorithm directly. In the code [7], the std of local position and rotation are fixed as, 0.1 m for position error std, and 0.01 rad for rotation error std, respectively. The two factors are jointly optimized by ceres solver. Since the pose graph is quite sparse, the authors keep a huge window to get accurate and globally drift-free pose estimation ( not implemented in the code). Although many nodes are optimized, the transformation between the local frame and the global frame is updated only by the global pose of the last node. As far as I am concerned, all the nodes should be taken into account to reduce the effect of possible coarse errors.

However, the **time synchronization** between the VIO and GPS measurements is not decently handled in the code, which may introduce significant pose error. Moreover, the integration of GPS and VIO is in a loosely-couple fashion; thus, the GPS observations are not used to correct the sensor errors of the IMU.

## VII. APPENDIX

### A. *Attitude Perturbation*

In most optimization problems, we calculate the Jacobian of residuals w.r.t. the variables at first. Then we use the output increment to update the state of the vehicle. So the way that we compute the Jacobian is relative to the way we update the state of body.

There are many attitude representations to parameterize the attitude of vehicles. There definitions and the conversions from one form to another can be found in [10, 18]. It is worth mentioning that the rotation vector and rotation matrix correspond to the Lie-algebra $so(3)$ and Lie-group $\mathrm{SO}(3)$, respectively.

For instance, the authors in [5] use the quaternion to represent the attitude of the vehicle, and the rotation matrix to represent the transformation between two coordinates. The authors define the update way of the vehicle attitude as (see “pose_local_parameterization.cpp” in the code)

$$\boldsymbol{q}_{b'}^{w} = \boldsymbol{q}_{b}^{w} \delta \boldsymbol{q}_{b'}^{b} \tag{71}$$

where, $\boldsymbol{q}_{b}^{w}$ is the vehicle attitude derived from the system dynamic eqation; $\delta \boldsymbol{q}_{b'}^{b}$ is the increment calculated by the nonlinear optimization. $\boldsymbol{q}_{b'}^{w}$ is the attitude after optimization. So the perturbation should be the attitude error in b-frame (from current $b$ to $b'$ ) when we perform perturbation analysis on the residual function. For example,

$$\begin{aligned} \mathbf{R}_{b}^{w}\left(\delta\boldsymbol{\theta}\right) &= \mathbf{R}_{b}^{w}\exp(\delta\boldsymbol{\theta}_{b'}^{b}) \approx \mathbf{R}_{b}^{w}\left(\mathbf{I} + \delta\boldsymbol{\theta}_{b'}^{b}\times\right) \\ \mathbf{R}_{w}^{b}\left(\delta\boldsymbol{\theta}\right) &= \exp(\delta\boldsymbol{\theta}_{b}^{b'})\mathbf{R}_{w}^{b} \approx \left(\mathbf{I} - \delta\boldsymbol{\theta}_{b'}^{b}\times\right)\mathbf{R}_{w}^{b} \end{aligned} \tag{72}$$

Where, $\mathbf{R}_{b}^{w}\left(\delta\boldsymbol{\theta}\right)$ and $\mathbf{R}_{w}^{b}\left(\delta\boldsymbol{\theta}\right)$ are the attitudes with perturbation; $\delta\boldsymbol{\theta}_{b'}^{b}$ is the rotation vector corresponding to the attitude perturbation.

### B. *The Jacobian of The IMU Model*

Given Eq. (71) and (72), we have

**1) $\mathbf{J}[0]$:**

$$\frac{\partial \delta \boldsymbol{\alpha}_{b_{k+1}}^{b_k}}{\partial \boldsymbol{p}_{b_k}^{w}} = \lim_{\delta \boldsymbol{p}_{b_k}^{w} \to 0} \frac{\mathbf{R}_w^{b_k}(\boldsymbol{p}_{b_{k+1}}^{w} - (\boldsymbol{p}_{b_k}^{w} + \delta \boldsymbol{p}_{b_k}^{w}) - \boldsymbol{v}_{b_k}^{w} \Delta t_k + \frac{1}{2}\boldsymbol{g}^{w} \Delta t_k{}^2) - \hat{\boldsymbol{\alpha}}_{b_{k+1}}^{b_k} - \delta \boldsymbol{\alpha}_{b_{k+1}}^{b_k}}{\delta \boldsymbol{p}_{b_k}^{w}} \tag{73}$$
$$= -\mathbf{R}_w^{b_k}$$

Perform the Lie-algebra left multiplication perturbation model [13], we have

$$\frac{\partial \delta \boldsymbol{\alpha}_{b_{k+1}}^{b_k}}{\partial \boldsymbol{q}_{b_k}^{w}} = \lim_{\delta \boldsymbol{\theta}_{b_k}^{b_k} \to 0} \frac{\exp(\delta \boldsymbol{\theta}_{b_k}^{b_k'})\mathbf{R}_w^{b_k}\left(\boldsymbol{p}_{b_{k+1}}^{w} - \boldsymbol{p}_{b_k}^{w} - \boldsymbol{v}_{b_k}^{w}\Delta t_k + \frac{1}{2}\boldsymbol{g}^{w}\Delta t_k{}^2\right) - \mathbf{R}_w^{b_k}\left(\boldsymbol{p}_{b_{k+1}}^{w} - \boldsymbol{p}_{b_k}^{w} - \boldsymbol{v}_{b_k}^{w}\Delta t_k + \frac{1}{2}\boldsymbol{g}^{w}\Delta t_k{}^2\right)}{\partial \boldsymbol{\theta}_{b_k}^{w}}$$
$$= \lim_{\delta \boldsymbol{\theta}_{b_k}^{bk} \to 0} \frac{\left(\left(\mathbf{I} - \delta \boldsymbol{\theta}_{b_k'}^{b_k} \times\right)\mathbf{R}_w^{b_k} - \mathbf{R}_w^{b_k}\right)\left(\boldsymbol{p}_{b_{k+1}}^{w} - \boldsymbol{p}_{b_k}^{w} - \boldsymbol{v}_{b_k}^{w}\Delta t_k + \frac{1}{2}\boldsymbol{g}^{w}\Delta t_k{}^2\right)}{\partial \boldsymbol{\theta}_{b_k}^{w}}$$
$$= \left[\mathbf{R}_w^{b_k}(\boldsymbol{p}_{b_{k+1}}^{w} - \boldsymbol{p}_{b_k}^{w} - \boldsymbol{v}_{b_k}^{w}\Delta t_k + \frac{1}{2}\boldsymbol{g}^{w}\Delta t_k{}^2)\right]\times \tag{74}$$

One thing that we must pay attention to is that the cross production does not confirm the associative law, namely

$$\mathbf{A}(\boldsymbol{p}\times)\mathbf{B} \neq (\mathbf{A}\boldsymbol{p})\times\mathbf{B} \tag{75}$$

From [10], we know

$$\delta \boldsymbol{q} \approx \begin{bmatrix} 1 \\ \frac{1}{2}\delta\boldsymbol{\theta} \end{bmatrix} \tag{76}$$

$$\boldsymbol{q}^* = \begin{bmatrix} q_w \\ -\boldsymbol{q}_v \end{bmatrix}$$
$$(\boldsymbol{p} \otimes \boldsymbol{q})^* = \boldsymbol{q}^* \otimes \boldsymbol{p}^* \tag{77}$$

For unit quaternion, we have

$$\boldsymbol{q}^{-1} = \boldsymbol{q}^* \tag{78}$$

$$L\left[\boldsymbol{q}^{-1}\right]_{3\times3} = R\left[\boldsymbol{q}\right]_{3\times3} \tag{79}$$

where, $L\left[\bullet\right]$ and $R\left[\bullet\right]$ are respectively the left- and right- quaternion-product matrices, $\left[\bullet\right]_{3\times3}$ is the $3\times3$ block at the right-bottom of the matrix.

$$\begin{aligned}
\frac{\partial\delta\boldsymbol{\theta}_{b_{k+1}}^{b_k}}{\partial\boldsymbol{q}_{b_k}^{w}} &= \lim_{\delta\theta_{b_k}^{w}\to 0}\frac{2\left[\left(\hat{\gamma}_{b_{k+1}}^{b_k}\right)^{-1}\otimes\left(\boldsymbol{q}_{b_k}^{w}\otimes\begin{bmatrix}1\\ \frac{1}{2}\delta\boldsymbol{\theta}_{b_k}^{w}\end{bmatrix}\right)^{-1}\otimes\boldsymbol{q}_{b_{k+1}}^{w}\right]_{xyz}-2\left[\left(\hat{\gamma}_{b_{k+1}}^{b_k}\right)^{-1}\otimes\left(\boldsymbol{q}_{b_k}^{w}\otimes\begin{bmatrix}1\\0\end{bmatrix}\right)^{-1}\otimes\boldsymbol{q}_{b_{k+1}}^{w}\right]_{xyz}}{\delta\boldsymbol{\theta}_{b_k}^{w}} \\
&\approx \lim_{\delta\theta_{b_k}^{w}\to 0}\frac{2\left[\left(\hat{\gamma}_{b_{k+1}}^{b_k}\right)^{-1}\otimes\begin{bmatrix}1\\ -\frac{1}{2}\delta\boldsymbol{\theta}_{b_k}^{w}\end{bmatrix}\otimes\left(\boldsymbol{q}_{b_k}^{w}\right)^{-1}\otimes\boldsymbol{q}_{b_{k+1}}^{w}\right]_{xyz}-2\left[\left(\hat{\gamma}_{b_{k+1}}^{b_k}\right)^{-1}\otimes\begin{bmatrix}1\\0\end{bmatrix}\otimes\left(\boldsymbol{q}_{b_k}^{w}\right)^{-1}\otimes\boldsymbol{q}_{b_{k+1}}^{w}\right]_{xyz}}{\delta\boldsymbol{\theta}_{b_k}^{w}} \\
&= \lim_{\delta\theta_{b_k}^{w}\to 0}\frac{2L\left[\left(\hat{\gamma}_{b_{k+1}}^{b_k}\right)^{-1}\right]R\left[\left(\boldsymbol{q}_{b_k}^{w}\right)^{-1}\otimes\boldsymbol{q}_{b_{k+1}}^{w}\right]\begin{bmatrix}0\\ -\frac{1}{2}\delta\boldsymbol{\theta}_{b_k}^{w}\end{bmatrix}}{\delta\boldsymbol{\theta}_{b_k}^{w}} \\
&= -L\left[\left(\hat{\gamma}_{b_{k+1}}^{b_k}\right)^{-1}\right]R\left[\left(\boldsymbol{q}_{b_k}^{w}\right)^{-1}\otimes\boldsymbol{q}_{b_{k+1}}^{w}\right]_{3\times3} \\
&= -L\left[\left(\boldsymbol{q}_{b_{k+1}}^{w}\right)^{-1}\otimes\boldsymbol{q}_{b_k}^{w}\right]R\left[\hat{\gamma}_{b_{k+1}}^{b_k}\right]_{3\times3}
\end{aligned}$$

(80)

As the same as Eq. (74), we have

$$\frac{\partial\delta\boldsymbol{\beta}_{b_{k+1}}^{b_k}}{\partial\boldsymbol{q}_{b_k}^{w}} = \left[\mathbf{R}_w^{b_k}(\boldsymbol{v}_{b_{k+1}}^{w}-\boldsymbol{v}_{b_k}^{w}+\boldsymbol{g}^{w}\Delta t_k)\right]\times \tag{81}$$

2) $\mathbf{J}\left[1\right]$:

$$\frac{\partial\delta\boldsymbol{\alpha}_{b_{k+1}}^{b_k}}{\partial\boldsymbol{v}_{b_k}^{w}} = \lim_{\delta\boldsymbol{v}_{b_k}^{w}\to 0}\frac{\mathbf{R}_w^{b_k}(\boldsymbol{p}_{b_{k+1}}^{w}-\boldsymbol{p}_{b_k}^{w}-\left(\boldsymbol{v}_{b_k}^{w}+\delta\boldsymbol{v}_{b_k}^{w}\right)\Delta t_k+\frac{1}{2}\boldsymbol{g}^{w}\Delta t_k{}^2)-\hat{\boldsymbol{\alpha}}_{b_{k+1}}^{b_k}-\delta\boldsymbol{\alpha}_{b_{k+1}}^{b_k}}{\delta\boldsymbol{v}_{b_k}^{w}} = -\mathbf{R}_w^{b_k}\Delta t_k$$

(82)

Given (20), we

$$
\begin{aligned}
\frac{\partial \delta \boldsymbol{\alpha}_{b_{k+1}}^{b_k}}{\partial \boldsymbol{b}_{ak}} &= \lim_{\delta \boldsymbol{b}_{ak} \to 0} \frac{-\left(\hat{\boldsymbol{\alpha}}_{b_{k+1}}^{b_k} + \mathbf{J}_{b_a}^{\alpha} \delta b_{a_k}\right) + \hat{\boldsymbol{\alpha}}_{b_{k+1}}^{b_k}}{\delta \boldsymbol{b}_{ak}} = -\mathbf{J}_{b_a}^{\alpha} \\
\frac{\partial \delta \boldsymbol{\alpha}_{b_{k+1}}^{b_k}}{\partial \boldsymbol{b}_{gk}} &= \lim_{\delta \boldsymbol{b}_{gk} \to 0} \frac{-\left(\hat{\boldsymbol{\alpha}}_{b_{k+1}}^{b_k} + \mathbf{J}_{b_g}^{\alpha} \delta b_{g_k}\right) + \hat{\boldsymbol{\alpha}}_{b_{k+1}}^{b_k}}{\delta \boldsymbol{b}_{gk}} = -\mathbf{J}_{b_g}^{\alpha}
\end{aligned}
\tag{83}
$$

$$
\begin{aligned}
\frac{\partial \delta \boldsymbol{\theta}_{b_{k+1}}^{b_k}}{\partial \boldsymbol{b}_{gk}} &= \lim_{\delta \boldsymbol{b}_{gk} \to 0} \frac{2\left[\left(\hat{\boldsymbol{\gamma}}_{b_{k+1}}^{b_k} \otimes \begin{bmatrix} 1 \\ \frac{1}{2}\mathbf{J}_{b_g}^{\gamma} \delta \boldsymbol{b}_{gk} \end{bmatrix}\right)^{-1} \otimes (\boldsymbol{q}_{b_k}^{w})^{-1} \otimes \boldsymbol{q}_{b_{k+1}}^{w}\right]_{xyz} - 2\left[\left(\hat{\boldsymbol{\gamma}}_{b_{k+1}}^{b_k}\right)^{-1} \otimes (\boldsymbol{q}_{b_k}^{w})^{-1} \otimes \boldsymbol{q}_{b_{k+1}}^{w}\right]_{xyz}}{\delta \boldsymbol{b}_{gk}} \\
&= \lim_{\delta \boldsymbol{b}_{gk} \to 0} \frac{2R\left[\left(\hat{\boldsymbol{\gamma}}_{b_{k+1}}^{b_k}\right)^{-1} \otimes (\boldsymbol{q}_{b_k}^{w})^{-1} \otimes \boldsymbol{q}_{b_{k+1}}^{w}\right] \begin{bmatrix} 0 \\ -\frac{1}{2}\mathbf{J}_{b_g}^{\gamma} \delta \boldsymbol{b}_{gk} \end{bmatrix}}{\delta \boldsymbol{b}_{gk}} \\
&= -R\left[\left(\hat{\boldsymbol{\gamma}}_{b_{k+1}}^{b_k}\right)^{-1} \otimes (\boldsymbol{q}_{b_k}^{w})^{-1} \otimes \boldsymbol{q}_{b_{k+1}}^{w}\right]_{3\times 3} \mathbf{J}_{b_g}^{\gamma}
\end{aligned}
\tag{84}
$$

As the same as Eq. (83) and (84), we have

$$
\begin{aligned}
\frac{\partial \delta \boldsymbol{\beta}_{b_{k+1}}^{b_k}}{\partial \boldsymbol{v}_{b_k}^{w}} &= -\mathbf{R}_{w}^{b_k} \\
\frac{\partial \delta \boldsymbol{\beta}_{b_{k+1}}^{b_k}}{\partial \boldsymbol{b}_{ak}} &= -\mathbf{J}_{b_a}^{\beta} \\
\frac{\partial \delta \boldsymbol{\beta}_{b_{k+1}}^{b_k}}{\partial \boldsymbol{b}_{gk}} &= -\mathbf{J}_{b_g}^{\beta}
\end{aligned}
\tag{85}
$$

$$
\begin{aligned}
\frac{\partial \delta \boldsymbol{b}_{a}}{\partial \boldsymbol{b}_{ak}} &= -\mathbf{I} \\
\frac{\partial \delta \boldsymbol{b}_{g}}{\partial \boldsymbol{b}_{gk}} &= -\mathbf{I}
\end{aligned}
\tag{86}
$$

3) $\mathbf{J}[2]$:

$$\frac{\partial \delta \boldsymbol{\alpha}_{b_{k+1}}^{b_k}}{\partial \boldsymbol{p}_{b_{k+1}}^{w}} = \lim_{\delta \boldsymbol{p}_{b_{k+1}}^{w} \to 0} \frac{\mathbf{R}_w^{b_k}(\boldsymbol{p}_{b_{k+1}}^{w} + \delta \boldsymbol{p}_{b_{k+1}}^{w} - \boldsymbol{p}_{b_k}^{w} - \boldsymbol{v}_{b_k}^{w} \Delta t_k + \frac{1}{2}\boldsymbol{g}^w \Delta t_k{}^2) - \hat{\boldsymbol{\alpha}}_{b_{k+1}}^{b_k} - \delta \boldsymbol{\alpha}_{b_{k+1}}^{b_k}}{\delta \boldsymbol{p}_{b_{k+1}}^{w}} = \mathbf{R}_w^{b_k} \tag{87}$$

$$\frac{\partial \delta \boldsymbol{\theta}_{b_{k+1}}^{b_k}}{\partial \boldsymbol{q}_{b_{k+1}}^{w}} = \lim_{\delta \boldsymbol{\theta}_{b_{k+1}}^{w} \to 0} \frac{2\left[\left(\hat{\boldsymbol{\gamma}}_{b_{k+1}}^{b_k}\right)^{-1} \otimes (\boldsymbol{q}_{b_k}^{w})^{-1} \otimes \left[\boldsymbol{q}_{b_{k+1}}^{w} \otimes \begin{bmatrix} 1 \\ \frac{1}{2}\delta \boldsymbol{\theta}_{b_{k+1}}^{w} \end{bmatrix}\right]\right]_{xyz} - 2\left[\left(\hat{\boldsymbol{\gamma}}_{b_{k+1}}^{b_k}\right)^{-1} \otimes (\boldsymbol{q}_{b_k}^{w})^{-1} \otimes \boldsymbol{q}_{b_{k+1}}^{w}\right]_{xyz}}{\delta \boldsymbol{\theta}_{b_{k+1}}^{w}} = L\left[\left(\hat{\boldsymbol{\gamma}}_{b_{k+1}}^{b_k}\right)^{-1} \otimes (\boldsymbol{q}_{b_k}^{w})^{-1} \otimes \boldsymbol{q}_{b_{k+1}}^{w}\right]_{3\times 3} \tag{88}$$

4) $\mathbf{J}[3]$:

$$\frac{\partial \delta \boldsymbol{\beta}_{b_{k+1}}^{b_k}}{\partial \boldsymbol{v}_{b_{k+1}}^{w}} = \mathbf{R}_w^{b_k} \quad \frac{\partial \delta \boldsymbol{b}_a}{\partial \boldsymbol{b}_{ak+1}} = -\mathbf{I} \quad \frac{\partial \delta \boldsymbol{b}_g}{\partial \boldsymbol{b}_{gk+1}} = -\mathbf{I} \tag{89}$$

*C. The Jacobian of The Vision Model*

Given Eq. (71) and (72), we have

**1)** $\mathbf{J}[0]$:

$$\frac{\partial \tilde{\boldsymbol{P}}_l^{c_j}}{\partial \boldsymbol{q}_{b_i}^{w}} = \lim_{\delta \boldsymbol{\theta}_{b_i'}^{b_i} \to 0} \frac{\mathbf{R}_b^c \left\{\mathbf{R}_w^{b_j}\left[\mathbf{R}_{b_i}^{w} \exp(\delta \boldsymbol{\theta}_{b_i'}^{b_i} \times)\left(\mathbf{R}_c^b \frac{1}{\lambda_l} \bar{\boldsymbol{P}}_l^{c_i} + \boldsymbol{P}_c^b\right) + \boldsymbol{P}_{b_i}^{w} - \boldsymbol{P}_{b_j}^{w}\right] - \boldsymbol{P}_c^b\right\} - \tilde{\boldsymbol{P}}_l^{c_j}}{\delta \boldsymbol{\theta}_{b_i'}^{b_i}} = \lim_{\delta \boldsymbol{\theta}_{b_i'}^{b_i} \to 0} \frac{\mathbf{R}_b^c \mathbf{R}_w^{b_j} \mathbf{R}_{b_i}^{w} (\mathbf{I} + \delta \boldsymbol{\theta}_{b_i'}^{b_i} \times)\left(\mathbf{R}_c^b \frac{1}{\lambda_l} \bar{\boldsymbol{P}}_l^{c_i} + \boldsymbol{P}_c^b\right)}{\delta \boldsymbol{\theta}_{b_i'}^{b_i}} = -\mathbf{R}_b^c \mathbf{R}_w^{b_j} \mathbf{R}_{b_i}^{w}\left[\left(\frac{1}{\lambda_l}\mathbf{R}_c^b \bar{\boldsymbol{P}}_l^{c_i} + \boldsymbol{P}_c^b\right)\times\right] \tag{90}$$

**2) $\mathbf{J}[1]$:**

$$\frac{\partial \tilde{\boldsymbol{P}}_l^{c_j}}{\partial \boldsymbol{q}_{b_j}^w} = \lim_{\delta\boldsymbol{\theta}_{b_j'}^{b_j} \to 0} \frac{\mathbf{R}_b^c \left\{ \exp\left(\delta\boldsymbol{\theta}_{b_j}^{b_j'} \times\right) \mathbf{R}_w^{b_j} \left[ \mathbf{R}_{b_i}^w \left( \mathbf{R}_c^b \frac{1}{\lambda_l} \bar{\boldsymbol{P}}_l^{c_i} + \boldsymbol{P}_c^b \right) + \boldsymbol{P}_{b_i}^w - \boldsymbol{P}_{b_j}^w \right] - \boldsymbol{P}_c^b \right\} - \tilde{\boldsymbol{P}}_l^{c_j}}{\delta\boldsymbol{\theta}_{b_j'}^{b_j}}$$

$$= \lim_{\delta\boldsymbol{\theta}_{b_j'}^{b_j} \to 0} \frac{\mathbf{R}_b^c \left\{ \left( \mathbf{I} - \delta\boldsymbol{\theta}_{b_j'}^{b_j} \times \right) \mathbf{R}_w^{b_j} \left[ \mathbf{R}_{b_i}^w \left( \mathbf{R}_c^b \frac{1}{\lambda_l} \bar{\boldsymbol{P}}_l^{c_i} + \boldsymbol{P}_c^b \right) + \boldsymbol{P}_{b_i}^w - \boldsymbol{P}_{b_j}^w \right] \right\}}{\delta\boldsymbol{\theta}_{b_j'}^{b_j}}$$

$$= \mathbf{R}_b^c \left\{ \left[ \mathbf{R}_w^{b_j} \left( \mathbf{R}_{b_i}^w \left( \mathbf{R}_c^b \frac{1}{\lambda_l} \bar{\boldsymbol{P}}_l^{c_i} \right) + \boldsymbol{P}_{b_i}^w - \boldsymbol{P}_{b_j}^w \right) \right] \times \right\} \tag{91}$$

**3) $\mathbf{J}[2]$:**

$$\frac{\partial \tilde{\boldsymbol{P}}_l^{c_j}}{\partial \boldsymbol{q}_c^b} = \lim_{\delta\boldsymbol{\theta}_{c'}^c \to 0} \frac{\exp\left(\delta\boldsymbol{\theta}_c^{c'}\right) \mathbf{R}_b^c \left\{ \mathbf{R}_w^{b_j} \left[ \mathbf{R}_{b_i}^w \left( \mathbf{R}_c^b \exp\left(\delta\boldsymbol{\theta}_{c'}^c\right) \frac{1}{\lambda_l} \bar{\boldsymbol{P}}_l^{c_i} + \boldsymbol{P}_c^b \right) + \boldsymbol{P}_{b_i}^w - \boldsymbol{P}_{b_j}^w \right] - \boldsymbol{P}_c^b \right\} - \tilde{\boldsymbol{P}}_l^{c_j}}{\delta\boldsymbol{\theta}_{c'}^c}$$

$$= \lim_{\delta\boldsymbol{\theta}_{c'}^c \to 0} \frac{\left( \mathbf{I} - \delta\boldsymbol{\theta}_{c'}^c \times \right) \mathbf{R}_b^c \left\{ \mathbf{R}_w^{b_j} \mathbf{R}_{b_i}^w \mathbf{R}_c^b \left( \mathbf{I} + \delta\boldsymbol{\theta}_{c'}^c \times \right) \frac{1}{\lambda_l} \bar{\boldsymbol{P}}_l^{c_i} + \mathbf{R}_w^{b_j} \mathbf{R}_{b_i}^w \boldsymbol{P}_c^b + \mathbf{R}_w^{b_j} \left( \boldsymbol{P}_{b_i}^w - \boldsymbol{P}_{b_j}^w \right) - \boldsymbol{P}_c^b \right\}}{\delta\boldsymbol{\theta}_{c'}^c}$$

$$\approx \left\{ \mathbf{R}_b^c \left[ \mathbf{R}_w^{b_j} \mathbf{R}_{b_i}^w \boldsymbol{P}_c^b + \mathbf{R}_w^{b_j} \left( \boldsymbol{P}_{b_i}^w - \boldsymbol{P}_{b_j}^w \right) - \boldsymbol{P}_c^b \right] \right\} \times +$$

$$\left( \mathbf{R}_b^c \mathbf{R}_w^{b_j} \mathbf{R}_{b_i}^w \mathbf{R}_c^b \frac{1}{\lambda_l} \bar{\boldsymbol{P}}_l^{c_i} \right) \times - \mathbf{R}_b^c \mathbf{R}_w^{b_j} \mathbf{R}_{b_i}^w \mathbf{R}_c^b \left( \frac{1}{\lambda_l} \bar{\boldsymbol{P}}_l^{c_i} \times \right) \tag{92}$$